\documentclass[acmsmall,authorversion,nonacm]{acmart}

\usepackage{color,soul}

\usepackage{enumitem}

\usepackage{booktabs}
\usepackage{multirow}
\usepackage{multicol}
\usepackage{makecell}

\usepackage{listings}
\usepackage{lstautogobble} 
\usepackage{color}         
\usepackage{zi4}          

\usepackage{soul}

\usepackage{algorithm}
\usepackage{algpseudocode}

\usepackage{tcolorbox}

\usepackage{rotating}

\usepackage{subcaption}

\usepackage{xspace}
\definecolor{codegreen}{rgb}{0,0.6,0}
\AtBeginDocument{%
  }

\begin{document}

\newboolean{showcomments}
\setboolean{showcomments}{true}
\ifthenelse{\boolean{showcomments}}
{\newcommand{\nb}[2] {
\fcolorbox{black}{gray!}{\bfseries\sffamily#1:}
{\sf\textit{#2}}
}
}
{\newcommand{\nb}[2]{}
}

\newcommand{\changed}[1]{{#1}}
\newcommand{\deleted}[1]{} %

\newcommand{\architecture}{\textsc{BiSupervised}\xspace} %
\newcommand{\imdb}{IMDB\xspace} %
\newcommand{\imgnet}{Imagenet\xspace} %
\newcommand{\squadvtwo}{SQuADv2\xspace} %

\title[Adopting Two Supervisors for Efficient Use of Large-Scale Remote DNN]{Adopting Two Supervisors for Efficient Use of Large-Scale Remote Deep Neural Networks}

\author{Michael Weiss}
\email{michael.weiss@usi.ch}
\orcid{0000-0002-8944-389X}
\affiliation{%
  \institution{Università della Svizzera italiana}
  \streetaddress{Via Giuseppe Buffi 13}
  \city{Lugano}
  \country{Switzerland}
  \postcode{6900}
}
\author{Paolo Tonella}
\email{paolo.tonella@usi.ch}
\orcid{0000-0003-3088-0339}
\affiliation{%
  \institution{Università della Svizzera italiana}
  \streetaddress{Via Giuseppe Buffi 13}
  \city{Lugano}
  \country{Switzerland}
  \postcode{6900}
}

\begin{abstract}
    Recent decades have seen the rise of large-scale Deep Neural Networks (DNNs) to achieve human-competitive performance in a variety of artificial intelligence tasks. 
    Often consisting of hundreds of millions, if not hundreds of billion parameters, these DNNs are too large to be deployed to, or efficiently run on resource-constrained devices such as mobile phones or IoT microcontrollers. 
    Systems relying on large-scale DNNs thus have to call the corresponding model over the network, leading to substantial costs for hosting and running the large-scale remote model, costs which are often charged on a per-use basis.
    In this paper, we propose \architecture, a novel architecture, where, before relying on a large remote DNN, a system attempts to make a prediction on a small-scale local model. A DNN supervisor monitors said prediction process and identifies easy inputs for which the local prediction can be trusted. For these inputs, the remote model does not have to be invoked, thus saving costs, while only marginally impacting the overall system accuracy.
    Our architecture furthermore foresees a second supervisor to monitor the remote predictions and identify inputs for which not even these can be trusted, allowing to raise an exception or run a fallback strategy instead.
    We evaluate the cost savings, and the ability to detect incorrectly predicted inputs on four diverse case studies: IMDB movie review sentiment classification, Github issue triaging, Imagenet image classification, and SQuADv2 free-text question answering.
\end{abstract}

\ccsdesc[500]{Computer systems organization~Embedded systems}
\ccsdesc[300]{Computer systems organization~Redundancy}
\ccsdesc{Computer systems organization~Robotics}
\ccsdesc[100]{Networks~Network reliability}

\keywords{datasets, neural networks, gaze detection, text tagging}

\maketitle

\section{Introduction}

In the last decades, we have seen a steady rise of \emph{Deep Neural Network} (DNN) components in modern software systems, so-called \emph{Deep Learning based Systems} (DLS):
For example, buyers are intelligently assisted when buying products online\footnote{E.g. \url{https://hellorep.ai/}},
customer service is performed and analyzed using bots~\cite{Chen2019Antprophet},
Github issues can be automatically triaged~\cite{kallis2019ticketTagger}
and medical diagnosis is made or supported by automated processes~\cite{ibrahim2021mlDiagnosis} (sometimes runnable even on a smartphone~\cite{yu2020malaria}).

To achieve ever better predictive accuracy, the size of the best-performing DNN models has increased enormously for many tasks, with some models now consisting of trillions of parameters.
Large models have to be hosted in server centers, amongst others due to their challenging hardware requirements. 
Client facing-applications thus have to make a choice between small, low-accuracy models running locally on the client's device (e.g. a browser or a mobile phone) or to offload their prediction tasks to a large model in the cloud, which amongst other disadvantages introduces latency and imposes a substantial financial cost to the software provider for hosting their model, or for using a third-party model.
For example, a single request to the popular GPT-3 model~\cite{brown2020GPT3}, hosted by the Non-Profit organization OpenAI for anyone to use, is billed  up to \$0.48\footnote{4000 tokens on a fine-tuned davinci-model (0.12\$ per 1000 tokens). See \url{beta.openai.com}.}. 

Even state-of-the-art models, given their typically intractably large input space and stochastic nature, will occasionally make wrong predictions. 
Here, typical machine learning evaluation metrics, such as high accuracy, are not sufficient to justify the inclusion of the model in many risk-averse settings.
Instead, software architects and software engineers have to defend against the imperfections of state-of-the-art DNN models and prevent system failure even when faced with rare, mispredicted inputs.
This led to the proposition and evaluation of various \emph{DNN supervisors} in the recent software engineering literature~\citep[e.g.][]{Weiss2021FailSafe, Stocco2020, Stocco2020Poster, Weiss2021-SA, Wang2020Dissector, Michelmore2018Autonomous, Catak2021AutonomousDriving}. A DNN supervisor is a software component implemented to monitor a DNN at runtime and to detect inputs for which the models' prediction cannot be trusted.
For such inputs, the system would default to a predefined domain-specific backup strategy (e.g. forwarding the request to a human operator) or to refuse to produce any output. 
Our work suggests using such supervisors not only to prevent system failure but also to reduce cost and latency in applications running on resource constraint devices.

Specifically, we propose \architecture, a novel software architecture that combines the advantages of cheap and fast local DNN with the advantages of using a highly accurate but costly remote DNN. Specifically, \architecture leverages a first supervisor to discriminate between inputs that can reliably be predicted locally on a small model, and only leverage the large remote model when the input is too challenging for the local model. A second supervisor is then used to filter out highly uncertain remote predictions.
Specifically, our contributions are the following:

\begin{description}[noitemsep]
\item [Architecture] We propose a novel architecture that combines small, local models with large remote ones, allowing us to fine-tune an optimal cost-reliability trade-off.
\item [Evaluation of Cost Savings] We empirically evaluate the trade-off between cost saving (in terms of the number of requests to the remote model) and system accuracy enabled by our architecture. Specifically, we evaluate if our architecture allows significant cost savings while maintaining a system-level accuracy at most marginally lower than the one of the remote model. We found that for all case studies, when reducing costs by 50\%, \architecture achieved a system-level accuracy which is barely lower than the one of the large-scale remote model. Moreover, in two case studies, \architecture even achieved an accuracy exceeding the one of the large-scale remote models, due to complementarities in the local and remote models capabilities.
\item [Evaluation of Supervised Performance] We assess \architecture in a supervised setting, i.e., including the detection of inputs for which even the remote DNN prediction is not reliable enough to be trusted and for which thus a system-specific healing strategy should be triggered. We compare the supervised performance of our system against that of a locally supervised model. We found that in the vast majority of the tested configurations, the supervised performance of \architecture is similar or exceeds the one of the locally supervised model.
\end{description}

\section{Background}

\subsection{The trend towards huge DNN.}
\label{sec:huge_dnns}

\paragraph{From 1 thousand to 1.6 trillion parameters in 33 years}
In 1989, \citet{lecun1989backpropagation} proposed a DNN achieving state-of-the-art classification performance on handwritten digits, marking one of the earliest real-world applications of a neural net trained end-to-end with backpropagation~\cite{karpathy2022Lecun}.
Consisting of only 1000 parameters, it is orders of magnitude smaller than what we use nowadays to achieve state-of-the-art (SOTA) performance on challenging tasks.
After 1989, we were still a long way from widespread adoption of DNNs for practical tasks: Training DNNs is a highly computation-intensive task and thus for a long time, faster classical methods of AI, such as Support Vector Machines continued to achieve better results on nontrivial tasks, still winning, for example, the image classification benchmark (\textsc{ILSVRC}), in 2010 and 2011~\cite{russakovsky2015imagenet}.
This changed with an innovation that would make the training of DNNs much faster, thus allowing for much larger DNNs:
Starting in 2010, the first successful application of plain backpropagation on a GPU achieved a speedup of 50 over CPUs~\cite{cirecsan2010gpu}, training networks of up to 12 million parameters.
Shortly thereafter, AlexNet, a 60 million parameter DNN~\cite{Krizhevsky2012Alexnet} won \textsc{ILSVRC2012}, marking a ``turning point for large-scale object recognition''~\cite{russakovsky2015imagenet}.
As of this writing, CoCa, the SOTA model for the same task consists of 2.1 billion parameters\footnote{According to \url{https://paperswithcode.com/sota/image-classification-on-imagenet}, accessed on July 21, 2022}~\cite{yu2022coca}.

In other domains, such as Natural Language Processing (NLP), we observe even larger dimensions: 
GPT-2 (1.5B parameters) in 2019~\cite{radford2019GPT2} and GPT-3 (175B parameters) in 2020~\cite{brown2020GPT3} led to global media coverage, due to their good performance in generating naturally-appearing text. 
Within the last two years, they have again been overtaken by even larger models, for example by \textsc{Megatron} (530B parameters, 2022)~\cite{smith2022megatron}.
To the best of our knowledge, the currently largest well-known DNN is \textsc{SwitchTransformer} with a parameter count of 1.6T~\cite{fedus2021switch}.

\paragraph{Cost of large models}
The large computational requirements of such large models when inferencing (i.e., predicting) present a major challenge to their adoption~\cite{mattina2020Cost}, as they require a large amount of memory, energy, and powerful CPUs or GPUs to make predictions.
This can make their deployment impractical if not impossible, especially regarding customer-facing, resource-constrained devices, and platforms such as web browsers, internet of things (IOT) microcontrollers, or mobile phones.
Large-scale, highly accurate DNNs are thus typically deployed in server centers, providing a remote inference service. 
There is a range of services facilitating such predictions, such as the \textsc{Huggingface Inference API}\footnote{https://huggingface.co/inference-api}, OpenAIs APIs to GPT-3\footnote{https://openai.com/api/} or Google's \textsc{Vertex AI}\footnote{https://cloud.google.com/vertex-ai/docs/start/introduction-unified-platform}.
Not only can already a single inference to a large model on such a service lead to high cost, 
the cost of such remote inferences also scales badly, as it increases quasi proportionally with the number of inference requests - providing essentially no economies of scale.

\subsection{Running DNNs in Resource-Constrained Enviroments}
\label{sec:resource_constrained_dnn}
\paragraph{Advantages of smaller but local models}
A substantially different approach is hence to run DNNs locally on the client devices, 
which constrains the size and complexity of the models, eventually leading to an inferior level of accuracy.\footnote{
    \architecture is not limited to classification problems, and other metrics to assess the quality of predictions than accuracy can thus be used for their assessment. For simplicity and readability, in this paper, we only refer to accuracy when discussing the quality of predictions. 
}
Such locally deployed models have a range of advantages, among which a much lower energy consumption, faster inference (due to the smaller model, but also as there is no need to do a network roundtrip), no dependency on a stable internet connection, and, as explained above, no per-inference cost to the software provider.

\paragraph{Means towards efficient models} 
There is thus much interest in making local models as efficient as possible, aiming to fit more parameters into the capacities of a resource-constrained system, or to optimize the performance of a model while keeping its number of parameters as low as possible. The following provides a non-comprehensive selection of some recent work and frameworks aiming to facilitate the use of DNN models in resource-constrained environments.
\begin{itemize}
    \item \emph{Small Model Architectures:} A range of works proposes specific DNN architectures or training procedures aiming to specifically facilitate deployment to resource-constrained devices. Examples are \textsc{TinyBERT}~\cite{jiao2019tinybert} and \textsc{MobileNet}~\cite{howard2017mobilenets, sandler2018mobilenetv2, howard2019mobilenetv3}.
    These models, which are comparably small for their tasks (NLP and Imagenet classification, respectively) are intended to be used as starting point to fine-tune a model to other specific use-cases, thus making them widely useful.
    
    \item \emph{Smaller Sibling Models:} Various recent releases of new, large-scale SOTA models were accompanied by a set of similar, but smaller models. The architectural choices and innovations  proposed to create the flagship large-scale SOTA models are applied also to a set of smaller models. 
    This allows users to make an accuracy vs model-size trade-off. Examples include \textsc{efficientnet}~\cite{tan2019efficientnet}, \textsc{OPT}~\cite{zhang2022opt} and \textsc{GPT-3}\footnote{Models are not publicly released, but access is allowed through an API.}~\cite{brown2020GPT3}.
    
    \item \emph{Model Tuning and Postprocessing:} Literature provides a range of techniques to make a given large DNN architecture fit for lower system requirements, allowing to deploy it on resource-constrained devices. 
    The most popular such approach is \emph{Quantization}, where a full-precision DNN is converted into a low-bitwidth integer version, thus reducing computational and memory requirements~\cite{yang2019quantization}. Another technique is DNN pruning where the number of weights is reduced in a guided way, aiming to only minimally reduce the model accuracy~\cite{liu2020autocompress}.
    
    \item \emph{Custom Frameworks:} A typical software stack used to build and train a DNN cannot easily be applied when deploying the DNN to an environment such as a mobile phone or web browser which, besides hardware constraints, also imposes constraints on the software stack. 
    To this end, dedicated frameworks have been proposed, among which the most popular is \textsc{tf-lite}\footnote{\url{https://www.tensorflow.org/lite}}, allowing to convert a standard tensorflow model into one that can be deployed on specific platforms, such as Android, iOS and Linux edge devices. A dedicated version, \textsc{tf-lite micro}  targets even particularly small embedded devices~\cite{david2020tflite}.
\end{itemize}

Naturally, none of these techniques allows us to deploy huge SOTA models in resource-constrained environments, thus not replacing the need for remote predictions when aiming for optimal prediction accuracy. 
They do however support and extend the possibility to deploy smaller models on local or edge devices in cases where suboptimal accuracy is acceptable.

\section{Related Work}

\subsection{Smart Offloading in Edge Computing}
The idea to delegate computationally intensive tasks from resource-constrained devices to the edge or to the cloud, denoted as \emph{offloading}, is not limited to DNNs. 
Recent work -- which is not specific to DNNs -- discusses the idea of intelligently deciding on the local device which tasks can be computed locally, and which inputs have to be forwarded to the remote computation API. 
Amongst other approaches, some work proposes to use a local DNN to categorize inputs to make said decision~\citep[Surveys ][]{saeik2021taskOffloadingSurvey, carvalho2020computationOffloadingUsingAI}. These architectures thus naturally show some high-level similarity with \architecture, specifically with the 1st level supervisor. 
However, our architecture, which is specific to DNN inference tasks, includes a 1st level supervisor that  is tightly integrated with the local model, allowing us to efficiently compute both a local prediction and the decision on whether a remote invocation is needed. Specifically, we leverage the possibility of quantifying the uncertainty of the DNN at inference time -- a possibility that does not generalize to arbitrary computations in local devices. To the best of our knowledge, ours is the first approach that supports smart offloading for DNNs running in a local, resource-constrained device.

\subsection{Network Supervision}
\label{sec:supervisors}

To account for the practically unavoidable imperfections in DNN predictions, the detection of inputs for which the corresponding prediction is likely to be wrong is a crucial component of Machine Learning safety~\cite{hendrycks2021unsolved}. 
This task, which we denote as \emph{supervision} (sometimes also refered to as \emph{monitoring}~\cite{hendrycks2021unsolved}, \emph{prediction validation}~\cite{Catak2022UncertaintyAware} or \emph{misbehavior prediction}~\cite{Stocco2020, Stocco2020Poster}) allows to ignore the DNNs prediction and instead run a fallback process (sometimes denoted healing process~\cite{Stocco2020}).

\subsubsection{Supervisors Used in Related Literature}

The literature offers a range of techniques that can be used for supervision.
In the following, we provide a brief overview.
While we make some recommendations in \autoref{sec:bisupervised_choice_of_firstlevel}, 
this overview aims to help the reader make an informed selection of supervisors to use in \architecture.

\begin{description}[noitemsep]
\item [Softmax-Likelihood Based] Most DNN classifiers use a \emph{Softmax}-normalized output layer: Let $sm$ denote the activations of such a softmax output layers, and let $C$ denote the set of classes. We have that $length(sm) = |C|$ and $\sum_{c\in C} sm_c = 1$. The models prediction is $argmax_{c\in C} sm_c$, with the assigned likelihood $max_{c\in C} sm_c$.\footnote{It is worth noting that $sm_c$ is not an appropriate approximation of $c$ actually being correct; it is well known that - if not deliberately calibrated - DNNs overestimate such likelihoods~\cite{Hendrycks2016Softmax}.}
Given a softmax output array, the simplest softmax-based supervisor, \emph{Vanilla Softmax} or \emph{MaxSoftmax}~\cite{Hendrycks2016Softmax}, simply trusts a prediction if its softmax value is above some threshold.
To that extent, other supervisors use functions of the entire softmax array, such as the prediction-confidence-score (the difference between the two highest likelihoods)~\cite{Zhang2020CharacterizingAdversarialDefects}, the entropy in the predicted likelihoods~\cite{Weiss2021UncertaintyWizard} or the Gini impurity in the predicted likelihoods~\cite{feng2020deepgini}. Outlier Exposure~\cite{Hendrycks2019OutlierExposure} is a technique to further strengthen the capabilities of Softmax-Likelihood-based supervisors during model training.

\item [Monte-Carlo Dropout (MC-Dropout)]  MC-Dropout is a simple way to implement a \emph{Bayesian Neural Network}, i.e., a network that allows predicting a distribution over outputs instead of just a single ``point'' prediction~\cite{jospin2022bayesian}. To this extent, it leverages multiple, randomized predictions to estimate the spread in the prediction and thus its the degree of uncertainty~\cite{gal2016dropout}. 
The randomness in the prediction process comes from enabling \emph{dropout} layers~\cite{srivastava2014dropout} at prediction time, i.e., layers that randomly drop some nodes' activations, which are typically used for regularization during DNN training.
Then, there is a range of techniques to quantify the models' uncertainty given the samples, such as the predictions \emph{variation ratio} or \emph{mutual information}. MC-Dropout has been widely used in the literature~\citep[e.g.][]{ Catak2021AutonomousDriving, Catak2022UncertaintyAware, Michelmore2018Autonomous, Zhang2020CharacterizingAdversarialDefects}.

\item [Ensembles] Similar to MC-Dropout, Ensembles~\cite{Lakshminarayanan2017Ensembles} estimate the models' uncertainty by measuring the agreement between sampled predictions for a given input. Here, however, the randomness in the prediction process comes from using a range of different DNNs, all sharing the same architecture, but trained using differently initialized weights and biases. Intuitively, inputs for which the different models disagree more are quantified with higher uncertainty. Here, the same quantifiers as the ones for MC-Dropout can be used.

\item [Surprise Adequacy] Instead of the approaches described above, which aim to quantify the models uncertainties, surprise adequacy (SA)~\cite{Kim2018SurpriseAdequacy} aims to quantify how \emph{surprising} (i.e., \emph{novel} or \emph{out-of-distribution}) an input is with respect to its activation traces in the supervised model.
There are different variants of surprise adequacy, such as \emph{Distance-based SA (DSA)}~\cite{Kim2018SurpriseAdequacy}, \emph{Likelihood-based SA (LSA)~\cite{Kim2018SurpriseAdequacy} and \emph{Mahalanobis-distance-based SA (MDSA)}~\cite{Kim2020ReducingLabellingCost, Kim2020EvaluatingSAforQA}}, some of which can be used in a multimodal setting~\cite{Kim2021MultiModalSA}, and we refer to the corresponding papers for a detailed description.

\item [Autoencoders] A more direct, black-box~\cite{Riccio2020} approach to supervision is taken by Autoencoder based supervision~\cite{Stocco2020, Stocco2020Poster, stocco2021confidenceDrivenAE, stocco2020towardsContinuously}. 
Autoencoders are DNNs aiming to first compress and then reconstruct an input. If trained on the same data as the supervised DNN, an autoencoder can detect out-of-distribution inputs at runtime: As such data was by definition not well represented in the training data, it is likely to have a higher reconstruction error.

\item [Other Supervisors] Alternative, less frequently used approaches include, \emph{Dissector}~\cite{Wang2020Dissector} which makes per-layer predictions and measures there disagreement, \emph{DeepRoad}~\cite{Zhang2018DeepRoad} a principal component analysis (PCA) based approach on the model's activation traces,  \emph{Learnt Assertions}~\cite{Lu2019LearntAssertions} which aims to automatically learn to recognize  inputs that are likely  to be mispredicted from the associated activations, and \emph{NIRVANA}~\cite{Catak2022UncertaintyAware} which combines multiple MC-Dropout quantifiers outputs in a support-vector machine (SVM).
\end{description}

\subsubsection{Comparison of Supervisors Regarding Computational Overhead}
In resource-constrained environments, the overhead imposed by a supervisor, i.e., the additional computation cost imposed by running the supervisor might be an important concern. 
Here, we see substantial differences between the supervisors: 
Clearly, Softmax-based approaches are the fastest approaches, as they do not require any additional predictions. Instead, the quantification of uncertainty from the softmax array requires just a few floating point operations. In the case of MaxSoftmax, there's actually just a single array value read required (the softmax likelihood for the predicted class), the associated overhead being clearly minimal.
MC-Dropout (and thus also NIRVANA) and Ensembles require substantially more time, as various predictions for the same input have to be collected. 
While in an existing tool~\cite{Weiss2021UncertaintyWizard}, we provide utilities to perform such predictions very efficiently, it still remains a show-blocker for very resource-constrained applications.

Autoencoders can also be considered quite fast and efficient, as there is only a single additional prediction (the autoencoder forward pass) required. Moreover, such additional prediction can actually be done at the same time or even before the prediction on the supervised model is made. Should the autoencoder-based supervisor thus decide that a prediction for a given input is not to be trusted, the prediction does not even have to be made. When originally proposed, the performance gain of using autoencoder-based supervision compared to the then state-of-the-arte DeepRoad was one of its major advantages~\cite{Stocco2020}.
The overhead of surprise adequacy much depends on its specific type: 
DSA and LSA suffer from major performance bottlenecks when large training sets are used. There are some ways to mitigate these problems, e.g. by using only a subsample of the training set and fast, vectorized implementations~\cite{Weiss2021-SA}, but even with such improvements, the use of these techniques in many real-world applications on devices such as microcontrollers remains prohibitive. 
In such domains, only the latest variant of SA, MDSA provides a low enough overhead~\cite{Weiss2022SimpleTechniques}.

\subsubsection{Comparison of Supervisors Regarding Supervision Performance}
Faced with the wide range of techniques applicable for supervision, a range of empirical studies has been performed comparing them~\cite{Berend2020CatsAreNotFish, Zhang2020CharacterizingAdversarialDefects, Weiss2021FailSafe, Weiss2022SimpleTechniques, Weiss2022Ambiguity, weiss2022stvr}.
Summarized, they all find that no single technique works as a dominant supervisor, i.e., no technique consistently outperforms all other approaches. 
Some studies highlight that ensembles provide often - but not always - a better supervision performance than all other techniques~\cite{Weiss2021FailSafe, weiss2022stvr, Ovadia2019} as they not only give an accurate uncertainty estimate, but often better prediction in the first place.
Softmax-based approaches also often provide good supervision performance
and in a recent study we found no statistical difference between the performance to detect misclassified inputs between the different variants of softmax-based supervisors. 
Thus the simple MaxSoftmax might be a good choice in many situations.
Surprise Adequacies, while generally performing a bit worse than Softmax-based supervisors~\cite{Weiss2022SimpleTechniques, Weiss2022Ambiguity}, have the advantage that they can be applied to any DNN, not just to classification models.
Lastly, autoencoders perform well as long as the training set is relatively sparse in the input feature space. Instead, with feature-rich datasets autoencoders start to generalize too well for supervision, as they learn to also reconstruct outliers consisting of the same features~\cite{Weiss2022Ambiguity, riccio2023validators}

\subsubsection{Comparison of Supervisors Regarding Robustness}
It is important that supervisors detect root causes of DNN failures, such as adversarial inputs, corrupted inputs, invalid inputs or ambiguous inputs. 
In a recent study, we compared the capabilities of various supervisors in detecting these specific root causes of uncertainty~\cite{Weiss2022Ambiguity}.
Summarized, we found that all supervisors show some weaknesses. 
E.g., surprise adequacy, which tends to work well in detecting corrupted, adversarial and invalid inputs, fails to detect in-distribution ambiguous inputs (i.e., inputs with only aleatoric, but no epistemic uncertainty). 
Softmax-based supervisors can in principle detect all root causes, but are are vulnerable to adversarial attacks, which typically aim to modify the values of the output layer and can thus easily be configured to create high-confidence but wrong predictions.

\section{\architecture: Our Proposed Architecture}

\begin{figure}
    \centering
    \includegraphics[width=\textwidth]{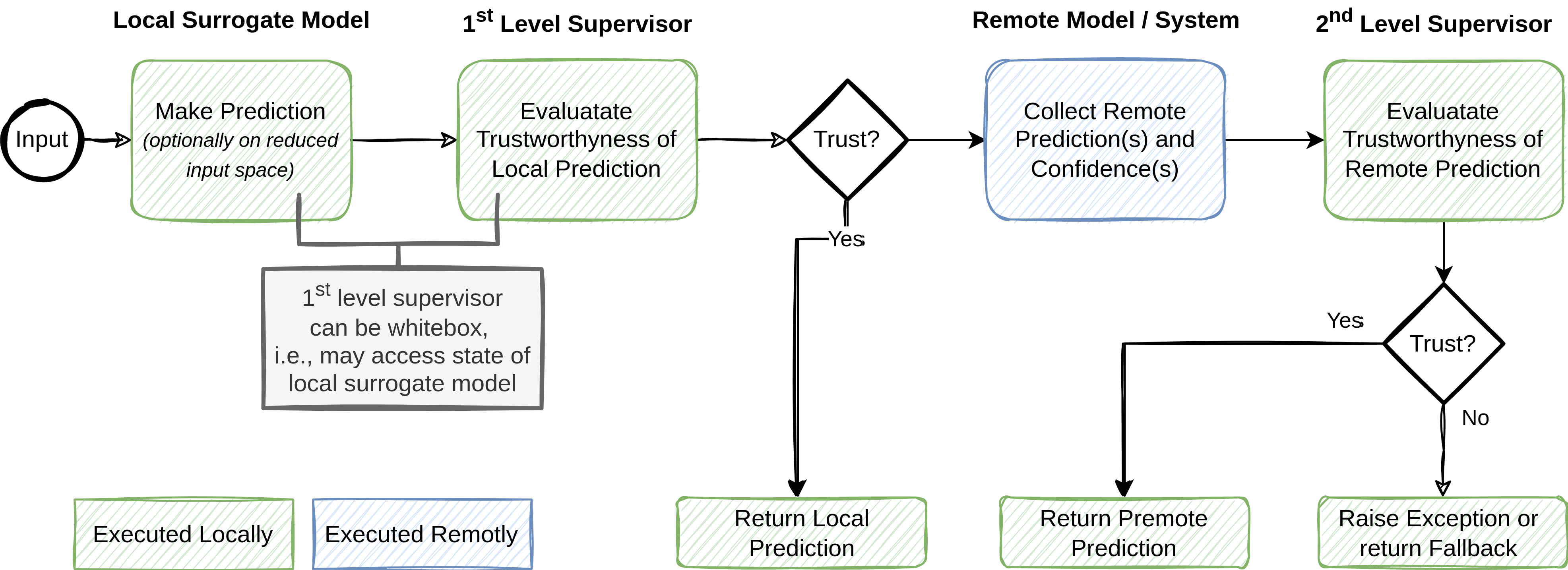}
    \caption{\architecture: Proposed Architecture for a Supervised Combination of Local and Remote Predictions}
    \label{fig:architecture}
\end{figure}

An overview of \architecture is illustrated in \autoref{fig:architecture}.
The goal of our architecture is as follows: (1) Attempt a prediction on a small, fast and cost-free \emph{local surrogate model} (short: \emph{local model)}. (2) Leverage a supervisor, denoted as \emph{1st level supervisor}, to identify inputs for which the local prediction cannot be trusted. (3) For those inputs, and only for those inputs, collect prediction and confidence from a large, highly accurate remote model. (4) Identify predictions for which not even the remote prediction can be trusted, based on the reported prediction and confidence in a \emph{2nd level supervisor} and raise a warning or return a default (fallback) value for those inputs.
Intuitively, this architecture attempts only to make costly remote predictions for \emph{difficult} inputs, where the simple local model is not sufficiently well trained.
The 2nd level supervisor furthermore allows handling inputs for which not even the highly accurate remote model can make a trustworthy prediction, such as invalid or ambiguous inputs.

Algorithm \ref{alg:bisupervised} outlines \architecture with threshold-based supervisors as pseudocode: On line 2-3 the local model makes a prediction and the 1st level supervisor quantifies the confidence of the prediction. To mitigate the computational effort required to execute lines 2-3, the corresponding input might be reduced in dimensionality (e.g. reducing the size of an image) to allow for smaller local models (line 1). 
If the confidence of this locally computed prediction is above some predefined threshold $t_l$, the prediction is returned (lines 4-6). Otherwise, the original input is sent to the remote prediction server, which returns the remote prediction amongst some metadata (e.g. prediction logits), from which the 2nd level supervisor can calculate a confidence (lines 7-8).
The remote prediction is returned, if the corresponding confidence is considered high enough. Otherwise, an exception is raised (lines 9-13).

A specific example highlighting how \architecture could be implemented and used in a practical application is given in Example \ref{ex:wildlife}

\begin{algorithm}
\caption{BiSupervised Algorithm}\label{alg:bisupervised}
\begin{algorithmic}[1]
\Require Bisupervised input $x$, local and remote confidence thresholds $t_l$ and $t_r$

\State $x_{small} \gets reduce\_dimensionality(x)$

\State $pred, metadata \gets local\_model.predict(x_{small})$

\State $conf \gets 1st\_level\_supervisor.calc\_confidence(metadata)$  

\If{$conf > t_l$}
\State \Return $pred$      
\EndIf

\State $pred, metadata \gets collect\_remote\_prediction(x)$  

\State $conf \gets 2nd\_level\_supervisor.calc\_confidence(metadata)$

\If{$conf > t_r$}
\State \Return $pred$  
\Else
\State \textbf{raise} Exception  \Comment{A high-confidence prediction was not found.}
\EndIf

\end{algorithmic}
\end{algorithm}

\begin{example}[Use of \architecture in a Wildlife Camera]
\label{ex:wildlife}

Let us consider a wildlife camera in a national park, installed to monitor the presence of wild animals and to report it to the park administration. 
Every 15 seconds, it takes an image that is classified according to the animal visible on the image, e.g. \emph{no-animal}, \emph{deer}, \emph{wolfe}, \emph{human}, etc.
Using \architecture, the classification works as follows:
First, the wildlife camera locally tries to classify the image. In doing so, it collects its full softmax output arrays, from which the 1st level supervisor can easily read the model's confidence~\cite{Hendrycks2016Softmax}.
If the confidence is high, the prediction is reported to the wildlife park's server. 
Otherwise, a remote model is leveraged:
the wildlife camera sends the image to a highly reliable 3rd party image classification service, such as IBM's Watson\footnote{\url{https://www.ibm.com/no-en/cloud/watson-visual-recognition/pricing}},
which returns its prediction and alongside the prediction logits.
The 2nd level supervisor then determines if the remote prediction can be trusted based on the remote model confidence (which can be determined based on the prediction logits). 
If the remote prediction is trustworthy, the prediction is reported to the  wildlife-park server. Otherwise, a warning is submitted to the park administration, alongside the unclassifiable image. 
The following scenarios may thus happen for different images:
\begin{itemize}
    \item \textbf{Nominal Image} Assume no animal stands in front of the camera. The local model confidently detects \emph{no-animal}, with low surprise adequacy. 
    No remote model invocation is needed, which thus saves money.
    \item \textbf{Rare Image} Assume there's fog, a condition which is not well represented in the local models' training set, thus leading to a non-trustworthy prediction of \emph{Beaver}. 
    The local supervisor detects the surprising input and thus ignores the local prediction, but forwards the image to the large-scale remote model, which generalizes better to such rare conditions and clearly detects a \emph{deer} in the fog. "\emph{Deer}" is thus reported to the park administration. 
    \item \textbf{Invalid Image} A storm caused the camera to be partially covered by mud, in a way that neither the local nor the remote model can make confident predictions, which both supervisors detect. A warning is sent to the park administration. With several such warnings reported in a short time, a park administrator double-checks the reported images and notifies a ranger to clean up the camera.  
\end{itemize}

\end{example}

In the following, we will explain each of the components in our architecture in more detail.

\subsection{Design of Local Surrogate Model}
The local model's objective is to make high-confidence predictions for as many inputs as possible in a very constrained environment:
Our architecture aims to be applicable to even tiny devices, such as microcontrollers, and as such, the design possibilities for the local model are limited, especially with respect to the number of model parameters. 
We refer to \autoref{sec:resource_constrained_dnn} for an overview of techniques to create such a model, but in the scope of our architecture, there's an additional measure that \emph{optionally} can be taken to allow a reduced model size, which we call \emph{Input Domain Reduction}:
Optionally, we may design a local model to handle a reduced input space instead of the original, possibly large, one: we can reduce the dimensionality of the input space, potentially losing some information, which may prevent  optimal predictions on a few inputs, but generally allows the local model to be much smaller in size.
For example, in NLP tasks, we may use a much smaller dictionary or clip the length of the input. 
For easy inputs, the information loss induced by the input domain reduction is unlikely to cause much uncertainty. For all other inputs, we  leverage a remote prediction on the large original input domain. 
E.g., in Example \ref{ex:wildlife}, we could train the local surrogate model on small grayscale images, and compress their inputs accordingly. Simple classifications, e.g. when no animal is around would still be possible with high confidence, but at the same time this would allow us to use a much smaller DNN.

\subsection{Design of 1st Level Supervisor}
\label{sec:bisupervised_choice_of_firstlevel}
In principle, our architecture is not tied to any specific supervision technique, thus all those listed in Section~\ref{sec:supervisors} could potentially be used as 1st level supervisor.
However, as the 1st level supervisor runs in the local, resource-constrained environment, one should make a choice that does not come with notable computational requirements during inference.
In our recent works~\cite{Weiss2021FailSafe, Weiss2022SimpleTechniques, Weiss2022Ambiguity}, we empirically compared a range of supervision techniques. 
Our results suggest the use of two specific techniques, as they both show a good supervision accuracy, while only causing negligible additional computational cost: 
For classification problems, where the local model makes softmax-normalized predictions, simply taking the highest softmax value (i.e., the softmax value of the predicted class)~\cite{Hendrycks2016Softmax} as a confidence score shows good practical supervision performance. 
For other models, without a softmax output, \emph{Mahalanobis-Distance based Surprise Adequacy (MDSA)}~\cite{Kim2020ReducingLabellingCost} reliably detects out-of-distribution inputs. 
We refer to \autoref{sec:supervisors} for a detailed discussion of different supervisors, their complementarities, weaknesses, strengths and overhead.

\subsection{Use of Remote Model or System and Design of 2nd-Level Supervisor}
Our architecture does not have any specific requirements on the architecture of the remote model.
Indeed, the remote model is considered a "black box" and may consist of more than just a DNN: in general, it might be a fully fledged, complex multi-component system (in this paper, we use the intuitive term \emph{remote model} for simplicity).
However, the 2nd level supervision can be heavily facilitated when the remote model is a DNN returning also its confidence (or related metrics, such as uncertainty) alongside its prediction. This may not be possible for all remote models, or may have to be specifically configured.
For example, using OpenAI's GPT-3, one can configure a request to return not just the sequence of tokens representing the model's maximum posterior, but a sequence of most likely options for each returned tokens, alongside their negative log-likelihood. 
This additional information about the model's prediction, which does not cause any additional cost, allows the straightforward implementation of a 2nd level supervisor.
Alternatively, a black-box or data-box~\cite{Riccio2020} supervisor (respectively, one with access to just the input/output of the remote model or also to the training data used to train the remote model) would have to be employed as 2nd level supervisor to select inputs that not even the remote model can handle. While this comes with yet another cost saving (the supervisor can run before the remote model, thus allowing to not collect remote predictions which will then be rejected), we do not recommend the use of such black- and data-box supervisors in this use case: They are likely to cause substantial additional load on the local resource-constrained platform, they typically do  not detect aleatoric uncertainty caused by ambiguity~\cite{Weiss2022Ambiguity} and lastly, in the case of data-box supervision, they require access to the training data of the remote model, which is typically not available or to too large to be easily processed.

\subsection{Superaccurate Performance}
The primary goal of our architecture is to allow a tradeoff between a cheap, low-accuracy model and an expensive, highly accurate remote model. 
As such, we expect the combined cost and accuracy to always lay between the ones of the local/remote model.
However, for what regards accuracy, our architecture may -- under specific circumstances -- even lead to accuracies above the one of the remote model, while still leading to reduced cost.
If the local and remote models' prediction capabilities are complementary (i.e., the set of inputs that the local model can correctly predict is not a subset of the inputs the remote model can predict), and the 1st level supervisor reliably trusts correct predictions by the local model, some inputs which would be incorrectly predicted by the remote model would be correctly predicted with our architecture.
With large complementarities, this could substantially increase the overall system accuracy, at a reduced cost. We denote such  improved overall system accuracy as \emph{superaccurate performance}.
However, aiming for complementarities between local and remote models might be infeasible, due to the constraints on the choice of models that can reasonably run on the local device:
In this paper, the main goal is to achieve cost-reduction with as little accuracy decrease as possible. 
Thus, while we might occasionally observe superaccurate performance empirically, its occurrence is more a (nice to have) side effect than an explicit design goal.

\subsection{Threshold Selection for Supervisors}
We strongly recommend the use of single-threshold supervisors in conjunction with \architecture.
Such supervisors are based on a single confidence (or uncertainty\footnote{We consider confidence and uncertainty as perfect complements~\cite{Riccio2020}, and for what regards DNN supervision, one is easily converted into the other by applying any order-reversing transformation to the value~\cite{Weiss2021FailSafe}}) score, and a configurable threshold. Inputs are trusted if and only if the confidence for said input is above a configurable threshold.
Modifying the threshold allows to gradually change the proportion of untrusted/trusted inputs. In \architecture, a carefully tuned threshold in the 1st level supervisor thus allows the selection of the desired percentage of inputs for which a remote prediction is collected (thus defining the cost savings). 
Similarly, a carefully tuned threshold in the 2nd level supervisor allows to regulate the percentage of inputs for which the system refuses to make a prediction. Viewing the 2nd level supervisor as a misprediction detector, changing the threshold allows to trade-off false positives vs false negatives.
Here, especially in domains which are not safety-critical, we can imagine that different users or different use-cases show varying degrees of risk-aversion. In a more risk-averse setting, the threshold can thus be tuned to allow more raised exceptions, possibly requiring manual intervention, while other use-cases may actually favor some mispredictions over downsides from a too frequently raised exception or too frequently called fallback (such as a decreased user experience due to overly frequent warnings).

\paragraph{Existing Techniques to Threshold Selection}
 Stocco et. al. \citep{Stocco2020} propose to fit the distribution supervision scores of nominal, correctly classified validation set inputs~\cite{Stocco2020}. 
This allows selecting a threshold with a specific false positive rate (i.e., expected percentage of false alarms), assuming that the distribution of inputs in production is the same as the one of the validation set.
The big advantage of this approach is that no dataset of outliers is needed and a nominal validation set is sufficient to calculate the threshold.
An alternative approach for threshold selection in test input validators was proposed by Dola et. al. (2021)~\cite{dola2021distribution}. They propose to choose a threshold by calculating supervision scores for both a nominal and an invalid dataset and then choose a threshold that separates these two distributions optimally. While they proposed these techniques for test data validations, it can clearly also be used for supervisors. 
However, here, it is worth noting that a realistic invalid dataset may not always be available. 
\architecture is agnostic to the way the threshold is set. We refer to Riccio et. al. (2023)~\cite{riccio2023validators} for a more in-depth comparison of the two approaches discussed here.

\paragraph{Runtime Configuration of Threshold}
Both approaches described above consider the threshold a constant, which is defined before deployment.
For \architecture, we suggest that in practice, a more dynamic way to set the threshold may be beneficial: 
Here, the threshold could be considered a  configuration variable that can be changed or tuned in production, for example using remote information~\cite{weinreich2003remote} or based specifically on the inputs observed on a specific user's local device.
Such dynamic threshold selection allows to quickly adapt the supervisor behavior, for example in cases where the input distribution changes, where more or less conservative supervision is desired (e.g. due to a change of cost in remote predictions), or where users specify their preferences in their user settings. To account for this diversity in our paper, we  evaluate our architecture using a range of different thresholds.

\subsection{Extensions of \architecture}
\architecture allows for some extensions of its architecture and use, which go beyond the scope of this paper but are worth mentioning.
First, edge nodes, where available, allow for an extension including a third supervised DNN on an edge node, with capabilities laying somewhere between the minimal local model and the huge remote model. As such, \architecture (which would effectively become \textsc{TriSupervised}), would invoke the model on the edge node if the local prediction cannot be trusted. Then, if also the edge model would face high uncertainty, the edge node could in turn forward the prediction to the remote model.

Second, we note that active learning could be easily integrated into \architecture: 
As the 1st level supervisor detects inputs for which the local model is (yet) insufficiently trained, collecting these inputs, labeling them (maybe even automatically using the remote model), and adding them to the local model's training set would allow for much more robust  generation of the next local model: In our recent work~\cite{Weiss2022SimpleTechniques}, we have shown that simple supervisors such as MaxSoftmax and MDSA, which we recommend in \architecture as 1st level supervisor, are also great techniques for such purpose.
Consequently, such an active learning setup would essentially allow a continuous improvement of the local model, thus further reducing the share of remote predictions needed in future versions of the software using \architecture.

\section{Empirical Evaluation of \architecture}

Our empirical evaluation aims to answer the following research questions:

\begin{description}[noitemsep]
\item [RQ1: Cost Savings] Using \architecture, can we reduce the number of remote predictions while only slightly impacting the overall prediction accuracy?
\item [RQ2: Supervision Performance] Provided with a given percentage of inputs to be used in remote predictions, how does \architecture perform in terms of \emph{supervised performance}~\cite{Weiss2021FailSafe}, compared to a standalone, supervised local model?
\end{description}

RQ1 is the main research question behind this work, as it aims at quantifying the trade off between cost saving and possibly reduced accuracy. This RQ ignores 2nd level supervision, as its focus is solely on the reduction of costly remote invocations, with no regard to the degree of confidence in the remotely produced prediction.

RQ2 aims at comparing the performance of a single local supervisor with that delivered by the two local/remote supervisors envisioned by \architecture. An ideal supervisor is one that blocks all and only inputs on which the model would otherwise exhibit poor accuracy. With RQ2 we check whether the proposed architecture, in addition to reducing the remote execution costs, can also improve the supervision performance.

\subsection{Assessment Metrics for RQ1}
All the supervisors used in our case studies work based on a single uncertainty score, which, if above a threshold, indicates that a prediction should not be trusted. 
We answer RQ1 in a threshold-agnostic way: Our results show the cost-accuracy tradeoff  across all possible thresholds, i.e., for any chosen proportion of inputs for which a remote prediction is collected.

\paragraph{Requests-Accuracy-Curve} Changing the threshold of the 1st level supervisor allows a trade-off between the percentage of inputs forwarded for remote prediction (fewer means lower cost) and high system accuracy (associated with many remote predictions). 
We visualize our results for RQ1 using a plot that we call \emph{Request-Accuracy-Curve} where the $x$ axis shows the percentage of inputs forwarded to the remote model and the $y$ axis shows the overall system accuracy (ignoring 2nd level supervision). 
The curve thus allows to quickly read cost savings and accuracy change when using our architecture with a specifically chosen 1st level supervisor threshold.
We will furthermore annotate selected points (e.g. knee points and superaccurate performances, if present) with the corresponding values, for even better readability of the plot.

\paragraph{Area Under Curve and Random Baseline}
The Request-Accuracy-Curve allows to visually compare against the random baseline, which is a straight line between the performance using all-local or all-remote predictions:
If successful, the curve produced by our architecture should be mostly above this line.
To quantitatively compare the random baseline against \architecture we measure the \emph{area under the curve of the Request-Accuracy-Curve} (AUC-RAC).
Specifically, we take the mean of all accuracies when setting the threshold to the value of each input, and normalize it on the accuracies observed for remote-only and local-only predictions. Let $\text{ACC}_{\text{SYS}}(r)$ denote the system performance for a given system and test set when a proportion of $r \in [0,1]$ of all inputs are forwarded to the remote model (thus, $r$ is a point in the $x$ axis on the Request-Accuracy-Curve) and  let $n$ denote the number of inputs in the test set. The AUC-RAC is then formally defined as:

\begin{equation}
    \frac{\frac{1}{n}\sum_{i=0}^n \text{ACC}_{\text{SYS}}(\frac{i}{n}) -  \text{ACC}_{\text{SYS}}(0)}{ \text{ACC}_{\text{SYS}}(1) -  \text{ACC}_{\text{SYS}}(0)}
\end{equation}

This assumes that the remote model exhibits a higher accuracy than the local model, i.e.,  $\text{ACC}_{\text{SYS}}(1) > \text{ACC}_{\text{SYS}}(0)$, which is a prerequisite for our experiments.
The expected AUC-RAC from random supervision is 0.5.
Note that, contrary to other AUCs, such as AUC-ROC, the proposed AUC-RAC can be below 0 in case of very bad supervision, or above 1 in case of very large superaccurate performance, due to normalization. Both of these are edge cases we do not expect to see frequently.

\subsection{Assessment Metrics for RQ2}
RQ2 considers the performance of \architecture in a supervised setting, i.e., including both supervisors to identify inputs for which neither local nor remote predictions can be trusted. 
To compare our architecture against a supervised local model, we use the following three metrics which we proposed and designed specifically to compare pairs of different supervised models:
\begin{itemize}
    \item \emph{Supervised Accuracy} ($\overline{ACC}$): The accuracy considering only the trusted predictions. Higher is better. Typically, the more conservative a supervisor, the higher $\overline{ACC}$.
    \item \emph{Acceptance Rate} ($\Delta$): The percentage of trusted inputs. Higher is better. Typically, the more conservative a supervisor, the lower $\Delta$.
    \item \emph{S-Beta Score} ($S_\beta$). The geometric mean between (normalized) supervised accuracy and acceptance rate. $\beta$ is the weight parameter in the geometric mean. 
    We will assess $S_{0.5}$, $S_1$, $S_2$.
\end{itemize}
We provide a detailed explanation of these metrics in ~\citet{Weiss2021FailSafe}. 

\paragraph{Threshold Selection}
Thus, for RQ2, we need to choose thresholds for both the 1st and 2nd level supervisor. For what regards the former, we use meaningful knee points identified in RQ1.
For what regards the latter, in line with related literature~\cite{Stocco2020, Weiss2021FailSafe, Catak2022UncertaintyAware}, we choose a threshold such that a targeted (overall) false positive rate (FPR) is reached.
Specifically, for each of the previously identified knee points, we will conduct three separate evaluations, in which we will tune the second-level supervisor's threshold such that FPRs of 0.01, 0.05, and 0.1 are met, respectively. As a baseline, we will assess the supervised performance of a system consisting of only the local model and supervisor using again thresholds chosen to lead to the same FPRs.
This also reflects our targeted use case where thresholds are runtime-tunable parameters, allowing them to be fine-tuned and adjusted as inputs (and shifts of inputs) are observed after deployment, while still allowing for a fair comparison with the baseline. 

\subsection{Case Studies}

We evaluate our architecture on four case studies.
They were chosen to reflect diverse problem types (text classification, image classification, and question answering) and to allow the inclusion of various different architectures as local and remote models.
An overview of the case studies is provided in \autoref{tab:case_studies}, and they are described  below. For implementation details, such as used hyperparameters and exact data pre-processing, we refer to our replication package, described in \autoref{sec:data_availabilty}.
\begin{table}[]
\caption{Overview of the case studies considered in our empirical evaluation}
\begin{tabular}{@{}lllllll@{}}
\toprule
Name                      & Domain                                               &        & Architecture                    & \#Params            & Metric              & Value \\ \midrule
\multirow{2}{*}{imdb}     & \multirow{2}{*}{Text Class.}     & Local  & Custom                          & 79K             & \multirow{2}{*}{Accuracy}       & 79.4\%      \\ \vspace{0.15cm}
                          &                                                      & Remote & GPT-3-curie \cite{brown2020GPT3}         & 13B * &                                 & 89.5\%      \\ 
\multirow{2}{*}{issues}   & \multirow{2}{*}{Text Class.}     & Local  & Custom                          & 223K               & \multirow{2}{*}{Micro Avg. F1} & 71.1\%           \\\vspace{0.15cm}
                          &                                                      & Remote & CatIss~\cite{izadi2022catiss} ** & 125M            &                                 & 82.3\%      \\
\multirow{2}{*}{imagenet} & \multirow{2}{*}{Image Class.} & Local  & MobilenetV3~\cite{howard2019mobilenetv3}           & 2.9M            & \multirow{2}{*}{Accuracy}       & 67.8\%      \\\vspace{0.15cm}
                          &                                                      & Remote & EfficientNetV2L~\cite{tan2021efficientnetv2}                & 119M            &                                 & 85.2\%      \\
\multirow{2}{*}{SQuADv2}                   & \multirow{2}{*}{Q\&A}                                  & Local  & Bert-Tiny~\cite{turc2019bertTiny} ***         & 6M              & \multirow{2}{*}{Exact Match}                      & 28.0\%           \\
                          &                                                      & Remote & GPT-3-davinci \cite{brown2020GPT3}        & 175B            &                                 & 30.8\%           \\ \midrule
 \multicolumn{7}{l}{\footnotesize * = estimated \cite{GP3ParamEstimate}, ** = finetuned roberta~\cite{Liu2019Roberta}, *** = finetuned}          \\ \bottomrule          
\end{tabular}
\label{tab:case_studies}
\end{table}

\subsubsection{\imdb Sentiment Classification (Text Classification)}

The \imdb sentiment classification benchmark~\cite{Maas2011imdb} is a common case study for NLP classification approaches.
Given an English-language movie review, a classifier is to decide whether the rating by the user is positive or negative. 
We use the following components for our architecture:
\begin{itemize}
    \item \textbf{Local Model:} For input domain reduction, we trim inputs to consist only of the first 100 words and use a small dictionary of only the 2000 most frequent words. We use a small model, consisting of one transformer block, a pooling, and two dense layers. For regularization, dropout layers have been added before the dense layers.
    The resulting local model has 78'558 parameters.
    \item \textbf{1st level supervisor:} As our local model is a softmax-based classifier, we can use vanilla softmax as the 1st-level supervisor.
    \item \textbf{Remote Model:} As remote model, we use a few-shot learning approach on the \emph{curie} model of the GPT-3~\cite{brown2020GPT3} family.
    With its estimated 13 billion parameters, it is more than 160'000 times larger than our local model. 
    We use a simple prompt, shown in Listing~\ref{lst:imdb-prompt}.
    This leads to an accuracy of 89.5\%, which is more than 9\% better than our local model.
    \item \textbf{2nd Level Supervisor:} Any valid response to our prompt consists of only one additional token. We configured our remote request to provide us with a list of 100 most likely tokens with their corresponding negative log-likelihood (easily transformable into likelihoods). We sum up the likelihood for known, hard-coded equivalent tokens (e.g. \textit{"Negative"}, \textit{"negative"} and \textit{"bad"}). This total likelihood is then used to determine whether or not the supervisor should reject the input.
\end{itemize}

We used the \imdb case study in a pilot experiment~\cite{Weiss2022CheapET3}, presented at the ESEC/FSE student research competition 2022. While showing encouraging results, motivating the writing of this paper, that pilot experiment answers only parts of RQ1, and considers a simpler architecture (not including a 2nd level supervisor), hence also not addressing RQ2.

\begin{lstlisting}[caption = {Few-shot GPT-3 Prompt for \imdb case study.}, label=lst:imdb-prompt]
Please classify the following movie reviews into the following sentiments: Negative, Positive.
---
Text: I hate this movie
Label: Negative
---
Text: I like this movie
Label: Positive
---
Text: < the review we want to classify >
Label:
\end{lstlisting}

\subsubsection{Github Issue Classifier (Text Classification)}
The Github Issue Classifier reflects a real-world use case: As new issues are reported to a github repository, maintainers typically triage the issues (are they a question, a bug report, an improvement request, etc.). \textsc{TicketTagger}~\cite{kallis2019ticketTagger} is an open source bot performing such triage by automatically labeling each input.
Run in a CI/CD environment, the ``local environment'' for such automated triager is typically quite resource-constraint.
Recently, \citet{kallis2022nlbse} released a corresponding benchmark dataset as part of the NLBSE'22 tool competition:
Given a few attributes of Github issues, amongst which title and description, issues have to be classified into one of the classes \textit{\{bug, enhancement, question\}}.
The dataset is unbalanced, heterogeneous (e.g., it contains outliers such as non-English issues), and contains some truly ambiguous inputs (i.e., issues for which the ground truth is not certain even to an experienced developer).
This makes this case study a particularly well-motivated setting for  RQ2, as unclear inputs should better remain unclassified and handled by a repository maintainer performing manual triaging.
Upon adopting \architecture, \textsc{TicketTagger} could assign a fallback label \emph{"manual triage required"} to inputs rejected by both supervisors.

To measure the model and system performance of this case study, we use a micro-average
F1-score, which takes into account the class imbalance in this case study, as described in the tool competition~\cite{kallis2022nlbse}. We then use the following components for our architecture:
\begin{itemize}
    \item \textbf{Local Model:} We use a similar transformer-based classifier as for \imdb, using the same type of layers, but with hyperparameters chosen to take the higher complexity of the problem into account (e.g., the hidden dense layer contains of 100 instead of 20 nodes) and with an increased dictionary size of 2000.
    As input we use the issue description concatenated to the issue title, truncated to the first 200 words.
    \item \textbf{Remote Model:} 
    As remote model, we use \textsc{CatIss}, the winning model of the NLBSE'22 tool competition~\cite{izadi2022catiss}.
    It is based on a finetuned Roberta model~\cite{Liu2019Roberta}. 
    As far as we know, this is the best available model for the issue classification dataset. We expect that by fine-tuning even larger Roberta instances, one may further improve the overall performance.
    \item \textbf{1st and 2nd level supervisor:}  
    We use vanilla softmax as 1st and 2nd level supervisors.
\end{itemize}

\subsubsection{\imgnet (Image Classification)}
\imgnet~\cite{russakovsky2015imagenet} is one of the most widely used image classification benchmarks in the literature.
Its images represent 1000 different classes, some of which very closely related to each other (e.g. 119 breeds of dogs~\cite{Eshed2020ImagenetNoveltyDetection}).
As such, it represents a challenging machine learning problem, where even large-scale state of the art models do not yet achieve accuracies above 91\%~\cite{yu2022coca}.
Provided that a wide range of \imgnet models are available through standard model repositories, and that such models are typically DNNs with a softmax output layer, choosing the components for our architecture is not difficult:
\begin{itemize}
    \item \textbf{Local Model:} We use \textsc{MobilenetV3}~\cite{howard2019mobilenetv3} as local model. \textsc{MobilenetV3} is the third and latest generation Mobilenet \imgnet model which is specifically designed to work in resource constrained environemnts.
    Various versions (differing in size) of \textsc{MobilenetV3} exist. We use `mobilenetv3\_small\_1.0\_224`, i.e., the model with 2.9 million parameters and an accuracy of 67.8\%\footnote{See \url{https://www.tensorflow.org/api_docs/python/tf/keras/applications/MobileNetV3Small}}
    \item \textbf{Remote Model:} We use \textsc{EfficientNetV2L}~\cite{tan2021efficientnetv2} as large-scale remote model. With its 119 million parameters and an accuracy of 85.2, it is the third-largest and most accurate \imgnet classifier available from Keras applications\footnote{\url{https://keras.io/api/applications/}, as of July 31st, 2022}.
    \item \textbf{1st and 2nd level supervisor:}  
    We use vanilla softmax as 1st and 2nd level supervisor.
\end{itemize}

\subsubsection{The Stanford Question Answering Dataset (Free-Text Question Answering)}

Our fourth case study considers \emph{question answering}, specifically the 2nd version of the \emph{The Stanford Question Answering Dataset} (SQuADv2)~\cite{rajpurkar2018know}.
Faced with a title and paragraph of English-language  Wikipedia, and a list of questions about the said paragraph, a model is supposed to provide a short answer, consisting of substrings of the paragraph (thus essentially free text), to each of the questions. 
SQuADv2 provides a set of correct answers for each question. For example, the question \textit{"What century did the Normans first gain their separate identity?"} lists  \textit{"10th century"}, \textit{"the first half of the 10th century"} and \textit{"10th"} as correct answers.\footnote{Technically, the answers dataset also contains the location in the paragraph, and answers occurring in the paragraph more than once are thus listed multiple times. For our evaluation, we only consider the answer, not its location.}
We use the \textit{Exact Match (EM)}~\cite{rajpurkar2018know} metric to assess the performance of our models and architecture, i.e., the percentage of predictions that exactly match one of the answers in the SQuADv2 ground truth answers.
In addition to the questions described above, SQuADv2 also contains \emph{unanswerable} questions, i.e. questions for which the answer cannot be found in the provided paragraph. They are ideal to test the models in a supervised setting. 
We ignore these unanswerable questions for RQ1 (which explicitly does not assess the 2nd level supervisor) but include them for RQ2 where they should be detected by the 2nd level supervisor.
Here, it is worth mentioning that much related literature using SQuADv2 considers an unanswerable input as correctly answered if the models' uncertainty is above the threshold (i.e., if the input is detected as unanswerable by the supervisor). 
We argue that this is not a realistic setting in practice (a refused prediction is certainly not equivalent to a correct one). 
Our assessment of RQ1 (EM only on answerable questions) and RQ2 ($\overline{ACC}$, $\Delta$ and $S_\beta$) are better suited. However, for this reason, the accuracies we find in RQ1, and the supervised accuracies we find in RQ2 are not consistent with the ones mentioned in the related literature for the same models.

\begin{itemize}
    \item \textbf{Local Model:} We use a fine-tuned version of \textsc{Bert-Tiny}~\cite{turc2019bertTiny}, a miniature version of the popular \emph{Bidirectional Encoder Representations from Transformers} (BERT) model~\cite{devlin2018bert}, well suited for deployment in resource-constrained environments.
    Fine-tuning to the SQuADv2 dataset has been performed by the Huggingface team\footnote{\url{https://huggingface.co/mrm8488/bert-tiny-5-finetuned-squadv2}}.
    \item \textbf{1st level supervisor:} Alongside the predicted tokens representing our local model answer, we collect an uncertainty score for each predicted token.
    We consider multiple ways suitable to reduce these scores  into a single one, which can be used for threshold-based supervision, with the product and the minimum being the most obvious candidates. 
    Indeed, we find that the literature mostly uses the product of these uncertainties. We do not consider this ideal for supervision: Given the very large amount of tokens, predicted probabilities, even for the simplest of inputs are likely to be much lower than one, which would naturally make the overall confidence in answers consisting of multiple tokens exponentially smaller than the confidence in single token predictions. 
    Thus, for our experiments, we use the \emph{minimum} of all predicted likelihoods of all predicted tokens. %
    \item \textbf{Remote Model:} As remote model, we use a one-shot learning approach on the \emph{davinci} model of the GPT-3~\cite{brown2020GPT3} family.
    With its 165 billion parameters, it is the largest model in the GPT-3 family and thus often understood as \emph{the} GPT-3 model. 
    Upon its first publication, the authors of GPT-3 assessed the model on the SQuADv2 dataset, where it showed good performance, in particular considering the one-shot (not fine-tuned) setting. 
    We build our prompt equivalent to the one built when assessing GPT-3. In short, the prompt is initiated with the title, followed by the paragraph and a single, answerable question with the correct answer (which is consequently removed from the test set) and finally the question for which we assess the answering capability (see example prompt in Figure G.28 of \citet{brown2020GPT3}).
    \item \textbf{2nd level supervisor:}  As explained in the \imdb case study, we can configure our requests to GPT-3 to collect a likelihood for every predicted token. We can then continue equivalently to the 1st level supervisor to reduce this array of likelihoods into a single confidence score, allowing threshold-based supervision.
\end{itemize}

For the Github issue classification and Imagenet case studies, we run the remote models on our highly-performing workstation used for our experimentation, as in our setting we can afford the required hardware, which is not the case in most client-based applications for which we propose \architecture.

\subsection{Results}

\def \racfigwidth {.6\linewidth}
\newcommand\racfigcaption[1]{Request-Accuracy Curve (RAC) for {#1}}

\subsubsection{Changes to Evaluation Setup}
The evaluation setup described above was peer-reviewed as a registered report before results were collected\footnote{\url{https://dl.acm.org/journal/tosem/registered-papers}}. The experiments have been conducted as anticipated, with the exception of the following two minor deviations:

\begin{enumerate}
    \item \textbf{Imdb Remote Supervisor:}
    We planned to use the sum of the likelihoods of equivalent tokens (e.g. \textit{"negative"}, \textit{"Negative"} and \textit{"bad"}) amongst the \textbf{100} most likely tokens. However, the OpenAI API restricts this information to the \textbf{5} most likely tokens\footnote{\url{https://beta.openai.com/docs/api-reference/completions/create\#completions/create-logprobs}} - which we thus use for our experiments.
    In general, we observe that subsequent equivalent tokens, in general, have a very low likelihood assigned. This is expected, as our prompt (see \autoref{lst:imdb-prompt}) suggests a specific spelling. Thus, we do not expect that the here-described change has any notable influence on the results.
    
    \item \textbf{SQuADv2 Prompt:}
    We anticipated using the same prompt as the authors of the original paper proposing GPT-3 when conducting their SQuADv2 assessment (see example prompt in Figure G.28 of \citet{brown2020GPT3}). During development, we found that using this prompt, GPT-3 would often give verbose, full-sentence answers. As such answers are not part of the SQuADv2 answer dataset, this could have led to a misrepresentative low EM (exact match) metric. 
    We thus extended the prompt stating explicitly that answers should be an \textit{"extraction from background, as short as possible"}. Here, background refers to the context given in the prompt, in line with the originally anticipated prompt.    
    This obvious fix is one which realistically any practitioner would have done and we thus decided to include it. 
    As GPT-3 is a closed source model, with unknown prompt-preprocessing, we do not know if the problem we observed was already present when Brown et. al.~\cite{brown2020GPT3} conducted their experiments. Given their decent results, we assume that the verbose sentences GPT-3 we observed when re-using their prompt might be caused by deliberate recent changes to the GPT-3 service (in most use-cases, verbose answers might be perceived as better), as otherwise, we would have expected Brown et. al.~\cite{brown2020GPT3} to adapt their prompt in a similar fashion as we did.
    While it is unclear what effect this change has on our results, it is likely that our changes actually lead to a stronger baseline for \architecture and overall a more ambitious experimental setup: With the anticipated higher EM of the standalone remote model, outperforming or performing similarly well - as measured in RQ1 -  becomes a more challenging task.
\end{enumerate}

\subsubsection{Results for RQ1: Cost Savings}

\begin{figure}
    \centering
    \includegraphics[width=\racfigwidth]{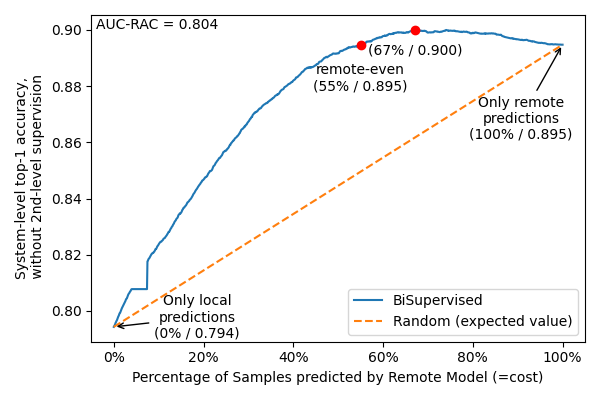}
    \caption{\racfigcaption{Imdb}}
    \label{fig:rac_imdb}
\end{figure}

\paragraph{Imdb} The RAC for the Imdb case study is presented in \autoref{fig:rac_imdb}. It shows a mild superaccurate performance:
At 68\% remote requests, the system-level accuracy (ignoring the 2nd-level supervisor) reaches its peak, which is mildly better than if only remote predictions are made. With 55\% remote requests, thus at 45\% remote prediction cost savings, \architecture achieves the same system-level performance as if only the remote model was used.

\paragraph{Github Issue Classification}

\def \squadracsubfigurewidth {0.5\linewidth}
\def \squadracimgwidth {.98\linewidth}

\begin{figure}
\centering
\begin{minipage}{\squadracsubfigurewidth}
  \centering
  \includegraphics[width=\squadracimgwidth]{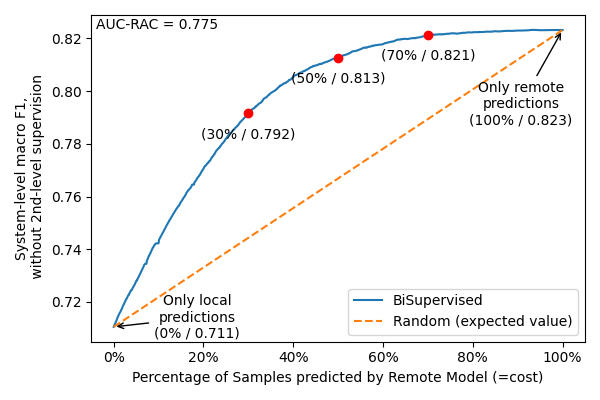}
  \caption{\racfigcaption{Issues}}
  \label{fig:rac_issues}
\end{minipage}%
\begin{minipage}{\squadracsubfigurewidth}
  \centering
  \includegraphics[width=\squadracimgwidth]{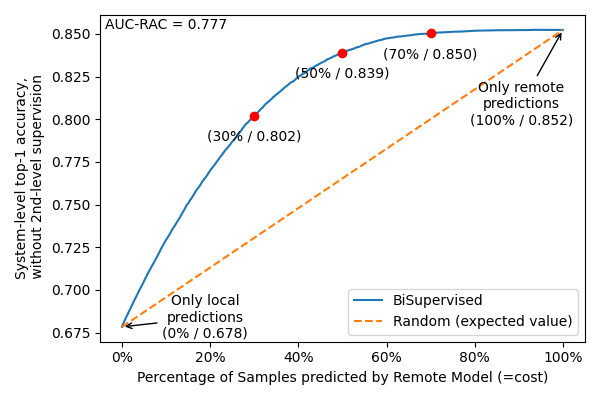}
  \caption{\racfigcaption{Imagenet}}
  \label{fig:rac_imagenet}
\end{minipage}
\end{figure}
The RAC for the GitHub Issue Classification case study is presented in \autoref{fig:rac_issues}. 
Here, \architecture allows to save a large part of remote-prediction costs while maintaining a system-level F1 similar to the one observed on the standalone remote model: With any threshold leading to at least 60\% remote predictions, the system-level F1 remains practically constant. \architecture thus allows saving 40\% remote prediction cost at essentially no negative impact on the system level F1.
Overall, the AUC-RAC of the plot is 0.775, thus clearly outperforming the random baseline (expected AUC-RAC 0.5).

\paragraph{Imagenet} 
The RAC for the Imagenet case study, presented in \autoref{fig:rac_imagenet} is similar to the one for the Github Issue Case study.
While there is no notable superaccurate performance, we observe again that \architecture allows saving 40\% remote prediction cost at essentially no negative impact on the system level accuracy.
At the same time, the steep increase of the RAC for small percentages of remote predictions highlights the potential \architecture has for applications that currently use \textsc{MobilenetV3} (the production-ready mobile-friendly model we use as a local model): With just a low percentage of predictions to a larger model (which may be placed on the edge or a remote server), the system-level accuracy is easily improved.

\paragraph{SQuADv2} 

\def \squadracsubfigurewidth {0.5\linewidth}
\def \squadracimgwidth {.98\linewidth}

\begin{figure}[t]
\centering
\begin{subfigure}{\squadracsubfigurewidth}
  \centering
  \includegraphics[width=\squadracimgwidth]{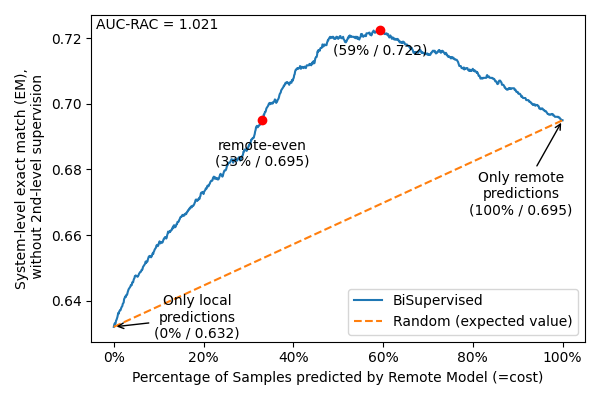}
  \caption{Only Valid (Answerable) Inputs}
  \label{fig:rac_squad_possible}
\end{subfigure}%
\begin{subfigure}{\squadracsubfigurewidth}
  \centering
  \includegraphics[width=\squadracimgwidth]{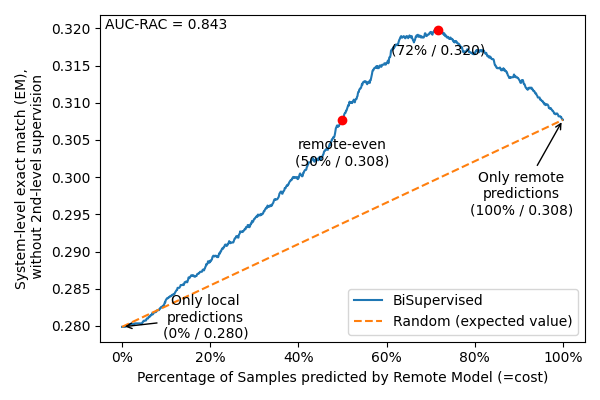}
  \caption{All Inputs}
  \label{fig:rac_squad_all}
\end{subfigure}%

\caption{
Request-Accuracy Curves (RAC) for SQuADv2 dataset. \architecture achieves superaccurate performances for both a clean dataset consisting of only valid (i.e., answerable) questions and for the full dataset.
}
\label{fig:rac_squadv2}
\end{figure}
As anticipated in the empirical setup, for what regards RQ1, the SQuADv2 dataset \emph{without} invalid inputs is of primary interest. The corresponding RAC is shown in \autoref{fig:rac_squad_possible}. 
Clearly, \architecture shows large superaccurate performance: At its peak, it achieves an EM of $0.722$, which is more than $.02$ better than the standalone remote model ($0.695$). At that point, only 59\% remote predictions are needed, thus also saving 41\% of remote prediction cost.
At only 38\% of the cost, \architecture achieves the same performance as the standalone remote model, clearly achieving the goal of cost savings.
Overall, \architecture achieves an AUC-RAC of $1.021$.
For completeness, we also calculated the RAC for the full SQuADv2 dataset, including all invalid questions, shown in \autoref{fig:rac_squad_all}. While these results are slightly inferior compared to the ones without invalid inputs, the plot still shows  superaccurate performance.

\begin{tcolorbox}[colback=cyan!5,colframe=cyan!75!black,title=Summary of RQ1 (Cost Savings)]
The results clearly show \textsc{BiSupervised}'s capability to save prediction cost while maintaining high accuracy:

\vspace{5px}
For two case studies, superaccurate performances were achieved, i.e., \architecture allows to get a better system-level prediction accuracy, at a much lower cost, compared to using only the large-scale remote model, or equivalent performance for 38\% or 55\% of the remote prediction cost, respectively.
In the other two case studies, where the models showed fewer complementarities, \architecture allowed saving roughly 40\% of remote prediction cost (dependent on the case study) while still achieving comparable system-level performance.

\vspace{5px}
In all case studies, \architecture achieved an AUC-RAC substantially above the  one associated with a random baseline.
\end{tcolorbox}

\subsubsection{Results for RQ2: Supervision Performance}
The results for RQ2 are presented in Tables \ref{tab:rq2_res_imdb}, \ref{tab:rq2_res_Issues}, \ref{tab:rq2_res_ImageNet}, \ref{tab:rq2_res_SQuADv2 (possible only)} and \ref{tab:rq2_res_SQuADv2 (all)}.
The RACs observed in RQ1 are smooth, which benefits our evaluation of RQ2 as the exact choice of 1st level supervisor becomes less critical: Small deviations in the threshold, i.e., in the percentage of inputs for which a remote prediction is made, do not have a major influence in the observed results. This allows us to choose thresholds for the 1st level supervisor for our evaluation of RQ2 in a generic way:
\begin{itemize}
    \item For case studies where we observe superaccurate performance (Imdb and SQuADv2) we use the thresholds leading to the highest performance, as well as the threshold corresponding to as few as possible remote predictions, leading to an overall performance equivalent to the one of the standalone remote model (denoted \emph{remote-even}).
    \item For case studies without superaccurate performance (Issue Classification and Imagenet) we choose thresholds for the 1st level supervisor corresponding to 30\%, 50\%, and 70\% remote predictions, which gives a good coverage of the observed curves, by considering both a point (30\%) with clearly reduced accuracy and one (70\%) with comparable overall accuracy, compared to a standalone remote model.
\end{itemize}

\paragraph{Imdb Case Study}

    \begin{table}
    \newcommand{\verti}[1]{\begin{tabular}{@{}c@{}}\rotatebox[origin=c]{90}{\parbox{1cm}{\centering #1}}\end{tabular}}
    \begin{tabular}{llcccccc}
\toprule
     &             & remote $\Delta$ &  $\Delta$ &  $\overline{obj}$ &  $S_{0.5}$ &  $S_1$ &  $S_2$ \\
FPR & remote requests &              &            &          &          &         &         \\
\midrule
\multirow{3}{*}{\verti{0.01}} & 0.00 (baseline) &          n/a &       0.98 &     0.80 &     0.83 &    0.88 &    0.94 \\
     & 0.55 (remote-even) &         0.96 &       0.98 &     0.91 &     0.92 &    0.94 &    0.96 \\
     & 0.68 (best) &         0.97 &       0.98 &     0.91 &     0.92 &    0.94 &    0.96 \\
\midrule \multirow{3}{*}{\verti{0.05}} & 0.00 (baseline) &          n/a &       0.93 &     0.82 &     0.83 &    0.87 &    0.90 \\
     & 0.55 (remote-even) &         0.86 &       0.92 &     0.92 &     0.92 &    0.92 &    0.92 \\
     & 0.68 (best) &         0.88 &       0.92 &     0.93 &     0.93 &    0.92 &    0.92 \\
\midrule \multirow{3}{*}{\verti{0.10}} & 0.00 (baseline) &          n/a &       0.86 &     0.83 &     0.83 &    0.85 &    0.86 \\
     & 0.55 (remote-even) &         0.76 &       0.87 &     0.93 &     0.91 &    0.90 &    0.88 \\
     & 0.68 (best) &          0.8 &       0.87 &     0.94 &     0.92 &    0.90 &    0.88 \\
\bottomrule
\end{tabular}

    \caption{System-Level Supervised Assessment for imdb}
    \label{tab:rq2_res_imdb} 
    \end{table}
    
The results for the Imdb base study are shown in \autoref{tab:rq2_res_imdb}. 
For all targeted false positive rates (FPR) and all tested local thresholds, i.e., for both the threshold leading to a remote-even system-level and best accuracy in RQ1, respectively, \architecture clearly outperforms the local baseline. The system-level acceptance rate ($\Delta$) is comparable to the local baseline in all cases, but the supervised accuracy ($\overline{obj}$) improves up to 11\% when using \architecture. 
Thus, for all weightings of $\Delta$ and $\overline{obj}$, their arithmetic mean, i.e., $S_{\beta}$ also outperforms the baseline.

\paragraph{Github Issues and Imagenet Case Studies}  

    \begin{table}
    \newcommand{\verti}[1]{\begin{tabular}{@{}c@{}}\rotatebox[origin=c]{90}{\parbox{1cm}{\centering #1}}\end{tabular}}
    \begin{tabular}{llcccccc}
\toprule
     &                & remote $\Delta$ &  $\Delta$ &  $\overline{obj}$ &  $S_{0.5}$ &  $S_1$ &  $S_2$ \\
FPR & remote requests &              &            &          &          &         &         \\
\midrule
\multirow{4}{*}{\verti{0.01}} & 0.00 (baseline) &          n/a &       0.98 &     0.72 &     0.76 &    0.83 &    0.91 \\
     & 0.30  &         0.94 &       0.98 &     0.80 &     0.83 &    0.88 &    0.94 \\
     & 0.50  &         0.96 &       0.98 &     0.82 &     0.85 &    0.89 &    0.94 \\
     & 0.70  &         0.97 &       0.98 &     0.83 &     0.86 &    0.90 &    0.95 \\
\midrule \multirow{4}{*}{\verti{0.05}} & 0.00 (baseline) &          n/a &       0.91 &     0.74 &     0.77 &    0.82 &    0.87 \\
     & 0.30  &         0.75 &       0.92 &     0.82 &     0.84 &    0.87 &    0.90 \\
     & 0.50  &         0.84 &       0.92 &     0.84 &     0.86 &    0.88 &    0.90 \\
     & 0.70  &         0.88 &       0.92 &     0.86 &     0.87 &    0.89 &    0.91 \\
\midrule \multirow{4}{*}{\verti{0.10}} & 0.00 (baseline) &          n/a &       0.84 &     0.77 &     0.78 &    0.80 &    0.83 \\
     & 0.30  &         0.55 &       0.86 &     0.82 &     0.83 &    0.84 &    0.86 \\
     & 0.50  &         0.71 &       0.85 &     0.86 &     0.86 &    0.86 &    0.86 \\
     & 0.70  &         0.79 &       0.85 &     0.88 &     0.87 &    0.87 &    0.86 \\
\bottomrule
\end{tabular}

    \caption{System-Level Supervised Assessment for Issues}
    \label{tab:rq2_res_Issues} 
    \end{table}

    \begin{table}
    \newcommand{\verti}[1]{\begin{tabular}{@{}c@{}}\rotatebox[origin=c]{90}{\parbox{1cm}{\centering #1}}\end{tabular}}
    \begin{tabular}{llcccccc}
\toprule
     &                & remote $\Delta$ &  $\Delta$ &  $\overline{obj}$ &  $S_{0.5}$ &  $S_1$ &  $S_2$ \\
FPR & remote requests &              &            &          &          &         &         \\
\midrule
\multirow{4}{*}{\verti{0.01}} & 0.00 (baseline) &          n/a &       0.96 &     0.70 &     0.74 &    0.81 &    0.89 \\
     & 0.30  &         0.93 &       0.98 &     0.81 &     0.84 &    0.89 &    0.94 \\
     & 0.50  &         0.96 &       0.98 &     0.85 &     0.87 &    0.91 &    0.95 \\
     & 0.70  &         0.97 &       0.98 &     0.86 &     0.88 &    0.92 &    0.95 \\
\midrule \multirow{4}{*}{\verti{0.05}} & 0.00 (baseline) &          n/a &       0.86 &     0.75 &     0.77 &    0.80 &    0.84 \\
     & 0.30  &         0.73 &       0.92 &     0.83 &     0.85 &    0.87 &    0.90 \\
     & 0.50  &         0.84 &       0.92 &     0.87 &     0.88 &    0.89 &    0.91 \\
     & 0.70  &         0.89 &       0.92 &     0.87 &     0.88 &    0.90 &    0.91 \\
\midrule \multirow{4}{*}{\verti{0.10}} & 0.00 (baseline) &          n/a &       0.77 &     0.79 &     0.79 &    0.78 &    0.78 \\
     & 0.30  &         0.53 &       0.86 &     0.84 &     0.84 &    0.85 &    0.86 \\
     & 0.50  &         0.69 &       0.85 &     0.89 &     0.88 &    0.87 &    0.85 \\
     & 0.70  &         0.78 &       0.85 &     0.90 &     0.89 &    0.87 &    0.86 \\
\bottomrule
\end{tabular}

    \caption{System-Level Supervised Assessment for ImageNet}
    \label{tab:rq2_res_ImageNet} 
    \end{table}
    
Same as for RQ1, the Github Issue and Imagenet case study show very similar results for RQ2 - presented in Tables \ref{tab:rq2_res_Issues} and \ref{tab:rq2_res_ImageNet} respectively.
For all tested thresholds and different $\beta$ in $S_\beta$, \architecture clearly outperforms the baseline. 
Compared to a standalone supervised local model, the use of the remote model in \architecture not only improves the system-level supervised accuracy $\overline{obj}$, but in most cases it leads also to a higher system-level acceptance rate $\Delta$.

\paragraph{SQuADv2 Case Study}

    \begin{table}
    \newcommand{\verti}[1]{\begin{tabular}{@{}c@{}}\rotatebox[origin=c]{90}{\parbox{1cm}{\centering #1}}\end{tabular}}
    \begin{tabular}{llcccccc}
\toprule
     &             & remote $\Delta$ &  $\Delta$ &  $\overline{obj}$ &  $S_{0.5}$ &  $S_1$ &  $S_2$ \\
FPR & remote requests &              &            &          &          &         &         \\
\midrule
\multirow{3}{*}{\verti{0.01}} & 0.00 (baseline) &          n/a &       0.97 &     0.64 &     0.69 &    0.77 &    0.88 \\
     & 0.33 (remote-even) &         0.94 &       0.98 &     0.70 &     0.74 &    0.82 &    0.91 \\
     & 0.58 (best) &         0.96 &       0.98 &     0.73 &     0.77 &    0.84 &    0.92 \\
\midrule \multirow{3}{*}{\verti{0.05}} & 0.00 (baseline) &          n/a &       0.90 &     0.67 &     0.70 &    0.77 &    0.84 \\
     & 0.33 (remote-even) &         0.79 &       0.93 &     0.71 &     0.75 &    0.81 &    0.88 \\
     & 0.58 (best) &         0.87 &       0.93 &     0.74 &     0.77 &    0.82 &    0.88 \\
\midrule \multirow{3}{*}{\verti{0.10}} & 0.00 (baseline) &          n/a &       0.81 &     0.70 &     0.72 &    0.75 &    0.79 \\
     & 0.33 (remote-even) &         0.62 &       0.87 &     0.72 &     0.74 &    0.79 &    0.84 \\
     & 0.58 (best) &         0.78 &       0.87 &     0.75 &     0.77 &    0.80 &    0.84 \\
\bottomrule
\end{tabular}

    \caption{System-Level Supervised Assessment for SQuADv2 (possible only)}
    \label{tab:rq2_res_SQuADv2 (possible only)} 
    \end{table}

    \begin{table}
    \newcommand{\verti}[1]{\begin{tabular}{@{}c@{}}\rotatebox[origin=c]{90}{\parbox{1cm}{\centering #1}}\end{tabular}}
    \begin{tabular}{llcccccc}
\toprule
     &             & remote $\Delta$ &  $\Delta$ &  $\overline{obj}$ &  $S_{0.5}$ &  $S_1$ &  $S_2$ \\
FPR & remote requests &              &            &          &          &         &         \\
\midrule
\multirow{3}{*}{\verti{0.01}} & 0.00 (baseline) &          n/a &       0.89 &     0.31 &     0.36 &    0.46 &    0.65 \\
     & 0.51 (remote-even) &         0.93 &       0.96 &     0.32 &     0.37 &    0.48 &    0.68 \\
     & 0.72 (best) &         0.94 &       0.96 &     0.33 &     0.38 &    0.49 &    0.69 \\
\midrule \multirow{3}{*}{\verti{0.05}} & 0.00 (baseline) &          n/a &       0.76 &     0.35 &     0.39 &    0.48 &    0.61 \\
     & 0.51 (remote-even) &         0.79 &       0.89 &     0.33 &     0.38 &    0.48 &    0.66 \\
     & 0.72 (best) &         0.86 &       0.90 &     0.34 &     0.39 &    0.49 &    0.68 \\
\midrule \multirow{3}{*}{\verti{0.10}} & 0.00 (baseline) &          n/a &       0.65 &     0.39 &     0.42 &    0.49 &    0.57 \\
     & 0.51 (remote-even) &         0.62 &       0.81 &     0.34 &     0.39 &    0.48 &    0.64 \\
     & 0.72 (best) &         0.78 &       0.84 &     0.34 &     0.39 &    0.49 &    0.65 \\
\bottomrule
\end{tabular}

    \caption{System-Level Supervised Assessment for SQuADv2 (all)}
    \label{tab:rq2_res_SQuADv2 (all)} 
    \end{table}
    
Considering only SQuADv2's valid answers (\autoref{tab:rq2_res_SQuADv2 (possible only)}), the results show a similar picture as for the Imdb case study in that \architecture consistently and clearly outperforms the baseline.
The picture is less clear, however, for the case which regards all inputs, i.e., including invalid inputs for which it is impossible for any model to give a correct answer. Here, for 63\% of the inputs, neither the local nor the remote model was able to make a correct prediction. 
The supervised performance of \architecture then depends on the targeted FPR: With a low FPR of $0.01$, \architecture still outperforms the local baseline as it succeeds in maintaining a higher acceptance rate $\Delta$ with similar or even mildly better system-level accuracy as the standalone local model.
However, as we increase the targeted FPR, especially with an FPR of $0.1$, the supervised accuracy using \architecture improves less than it does for the local model while maintaining a much higher system-level acceptance rate $\Delta$. 
Presumably, this is caused by the fact that we achieve a given target FPR by manipulating only the remote supervisor's threshold while keeping the 1st level supervisor's threshold constant: As such, we remove some correct remote predictions (quickly reaching the targeted FPR), while the wrong but accepted local predictions negatively impact $\overline{obj}$. Thus, in this case study, which of the approaches performs better in a supervised setting depends on the risk-adverseness of the user, i.e., the $\beta$ chosen in $S_\beta$: While \architecture, in the configuration tested by us, for FPR 0.1, outperforms the baseline for $S_2$, it fails to beat the baseline for \textbf{$S_{0.5}$}.
Overall, SQuADv2 (including invalid inputs) is the only case study with results that are not conclusively in favor of \architecture, while still showing  a positive tendency:
Out of 18 assessed settings (target FPR, $S_{\beta}$ and 1st level supervisor threshold), \architecture shows a result outperforming the baseline in 13 cases; it has inferior $S_{\beta}$ scores in 5 cases.

\begin{tcolorbox}[colback=cyan!5,colframe=cyan!75!black,title=Summary of RQ2 (Supervision Performance)]
The supervised performance achieved using \architecture was higher than the one of our baseline, the standalone supervised local model, for all case studies in the majority of configurations.
This is primarily caused by the higher system accuracy, paired with the capability of the two-level supervision process to detect generally difficult inputs.
\end{tcolorbox}

\subsection{Threats}

\paragraph{Internal validity threats:} Two case studies, namely \imdb and SQuADv2, use GPT-3 as  remote model, which was trained on a large (unknown) set of texts collected using web crawling. It is possible that these texts may contain entries from our test set (this is sometimes called \textit{test set pollution}). First, we deem this unlikely and not different from a realistic usage of a large-scale remote model such as GPT-3. Thus, it is not expected to impact the validity of our result w.r.t. a real-world usage of \architecture. Second, for SQuADv2, on which GPT-3 was originally assessed, the authors measured the impact of test set pollution and found it to not have any major influence on the observed capabilities~\cite{brown2020GPT3}.

Motivated by our use case, in particular the possibility of runtime adaptive configuration of supervisor thresholds, we did not choose the thresholds for our evaluation before running the experiments. In RQ1, we perform a threshold-independent analysis (AUC-RAC) \emph{and} we also plot the curves. Smooth curves indicate that the selection of a threshold in practice is easy (no extreme local optima). Indeed, the curves we observed in RQ1 show no big fluctuations and are comparably smooth. Regarding RQ2, where a threshold independent metric is not possible, we chose a wide range of different thresholds for both supervisors, based on the findings of RQ1, and fixed FPR for the 2nd level supervisor. The likelihood that a majority of them would end up in small local optima (which would invalidate our results) is thus negligible.

\paragraph{Construct validity threats:} Quantification of the cost-accuracy trade off and of the supervisors' performance is not straightforward and no consolidated metrics exist for such task. We defined the AUC-RAC metric for the former quantification, as it provides a threshold independent quantification suitable for visual display as a request-accuracy-curve that can be easily interpreted by humans. For the latter we adopted the S-Beta score, as proposed in recent work about supervisor assessment~\cite{Weiss2021FailSafe}.

\paragraph{External validity threats:} We considered only four case studies. To choose models representative of a variety of application domains, we included models designed for text classification, image classification and question answering. However, the validity of our findings beyond the considered case studies should be assessed empirically by replicating our study on different domains/models.

\subsection{Advantages beyond Cost Savings}
\label{sec:other_advantages}
Besides the cost savings achieved by our architecture, making fewer remote predictions has a range of other advantages. Those are however very dependent on the used infrastructure, and we thus discuss them conceptually instead of empirically.

\paragraph{Lower Latency}
Latency of distributed systems, such as our architecture, depends on many factors such as network speed, bandwidth, used hardware, and system load.
We can reasonably assume that requests over a network are time-wise quite costly and that large-scale DNNs take more time to make a prediction than small-scale DNNs.
While the latter can, to some extent, be mitigated by using very fast hardware for the large-scale model, the former is an unavoidable truth using today's networks.
In an abstract model of latency in \architecture, let $t_l$ be the average time required to locally preprocess the input, make a prediction on the local model and run the first-level supervisor. Similarly let $t_r$ denote the average time required to make a remote prediction (including time for network communication, and supervision of the remote prediction). Let $r$ be the rate of inputs for which \architecture triggers a remote prediction.
\architecture then shows, on average, lower latency if
\begin{equation}
        \label{eq:latency}
        t_r > t_l + rt_r \\
\end{equation}
This is true whenever $(1-r)t_r > t_l$.

\begin{table}[ht]
\footnotesize
\centering
\begin{tabular}{llllll}
\toprule
                         Case Study & \thead{Local\\Only} & \thead{Remote\\Only} & \thead{Break Even\\(*)} &                                         \thead{Eval Points\\(vs. Remote Only)} &                                                                                      Remarks \\
\midrule
                    \makecell{imdb} &               0.05s &                0.32s &      \makecell{85.68\%} &                        \makecell{55\%: 0.23s (-30.7\%)\\67\%: 0.26s (-18.5\%)} &                                                                   \vspace{5pt}\makecell[l]{} \\
                  \makecell{Issues} &               0.09s &                1.08s &      \makecell{91.53\%} & \makecell{30\%: 0.41s (-61.5\%)\\50\%: 0.63s (-41.5\%)\\70\%: 0.84s (-21.5\%)} &                                     \vspace{5pt}\makecell[l]{Remote model emulated locally.} \\
                \makecell{ImageNet} &               0.02s &                0.68s &      \makecell{96.35\%} & \makecell{30\%: 0.23s (-66.3\%)\\50\%: 0.36s (-46.3\%)\\70\%: 0.50s (-26.3\%)} & \vspace{5pt}\makecell[l]{Remote model emulated locally.\\Both models use batched inference.} \\
\makecell{SQuADv2\\(possible only)} &               0.02s &                0.71s &      \makecell{97.42\%} &                        \makecell{33\%: 0.25s (-64.4\%)\\59\%: 0.44s (-38.1\%)} &                                                                   \vspace{5pt}\makecell[l]{} \\
          \makecell{SQuADv2\\(all)} &               0.02s &                0.74s &      \makecell{97.53\%} &                        \makecell{49\%: 0.39s (-47.6\%)\\71\%: 0.55s (-25.9\%)} &                                                                               \makecell[l]{} \\
\midrule

\multicolumn{6}{l}{\makecell{(*) Break even point: the fraction of remote calls 
at which the \architecture latency equals the latency\\ of the remote-only approach.
For any 1st level supervisor threshold leading to fewer remote predictions,\\
\architecture is faster than using the standalone remote model. Hence, higher is better.}}\\
\bottomrule
\end{tabular}

\caption{Average Prediction and Supervision Latency per Input}
\label{tab:times}
\end{table}

We measure the latencies induced from using local and remote models, which allows us to quantify potential latency improvements from using \architecture over using the remote model. All measurements include all required computation, including data preprocessing, inference, supervision, and network requests, where applicable.
To collect our results, we used a workstation equipped with an AMD Ryzen Threadripper 1920X CPU and a wired internet connection. No GPU was used. Clearly, the measured times highly depend on this configuration, as well as additional influences such as remote server load. Thus, our reported times should be seen as an illustration, without a claim of generalizability.
Here it is worth noting that, while we present the time measurements for all case studies, they are most meaningful for the Imdb and SQuADv2 case studies, which use a truly remote model accessed through the OpenAI API. 
For Github Issue classification and Imagenet, the large scale ``remote'' model was emulated locally, thus  leading to no networking overhead, but also not leveraging dedicated hardware. We still report the latter results for completeness.

For all case studies, our measurements confirm the huge potential \architecture shows with regards to latency: With a 1st-level supervisor configured such that less than 85\% of requests are forwarded to the remote model, the average latency is lower than the baseline which uses only a remote model, thus fully compensating the additional overhead induced by the local inference and supervision.
For Imagenet and SQuADv2 case studies, this break-even point  lays even above 96\%.
Consequently, all evaluation points we discussed in RQ1 and RQ2 would reduce average latency in practice. 
Thus, \architecture not only saves cost, but it also reduces the average inference time, all while increasing robustness in a supervised setting and keeping the same accuracy or, in some cases, even increasing the system level accuracy.

\paragraph{Lower Energy Consumption}
Large models, naturally, require more energy to make predictions than small models, due to the substantially higher number of arithmetic operations to perform to make a single prediction. 
In addition, sending the request to a remote server also causes an energy consumption overhead.
Thus, similar to the latency discussed above, we can expect the small overhead introduced by making a local inference and 1st level supervision (which are designed to be energy efficient), to be amortized by the lower number of remote requests even with a relatively low fraction of exclusively-local predictions.

\section{Future Work}
At the moment, \architecture assumes that for all inputs where the local prediction is highly uncertain, the remote architecture potentially has a lower uncertainty. 
However, this may not always be the case: In some domains, \emph{aleatoric uncertainty}, i.e., the uncertainty inherent in the input for which a prediction has to be made (such as noise or ambiguity) is a major cause of wrong predictions~\cite{kendall2017uncertainties}. 
Here, we notice that only in cases where the local model exhibits ~\emph{epistemic uncertainty}, such as uncertainty caused by insufficient training data or model architecture, forwarding the input to a much larger, thus presumably more capable DNN is justified.
For future work, we thus propose to attempt to discriminate between aleatoric and epistemic uncertainty in the local supervisor and only make remote predictions for inputs that show a high level of epistemic uncertainty, as opposed to a high level of general uncertainty.

We conducted our experiments under nominal data, i.e., data that is mostly in-distribution of the training datasets. Reproduction of our results using outlier data to assess \architecture is needed to assess its robustness.

\section{Conclusion}

In this paper, we proposed \architecture, a novel architecture that addresses the challenge of using large-scale Deep Neural Networks (DNNs) from resource-constrained devices -- a problem that is often solved by using costly and high-latency prediction on a remote server. Our proposed architecture combines a small-scale local model with a large-scale remote model, and includes two supervisors that monitor the prediction process, identifying inputs for which the local prediction can be trusted, and those that require the remote model to be invoked.

We evaluated the effectiveness of \architecture on four diverse case studies, including Imdb movie review sentiment classification, Github issue triaging, Imagenet image classification, and SQuADv2 free-text question answering. Our results demonstrate that \architecture can achieve superaccurate performance for Imdb and SQuADv2, i.e., the accuracy achieved by \architecture exceeds the one from the large-scale remote model. For the other case studies, the remote inference cost can be reduced by up to 50\% while maintaining a comparable system-level accuracy. 

To assess the ability of \architecture in a supervised setting, i.e., when identifying inputs for which not even the remote prediction can be trusted, we compared it against the standalone supervised local model and found that \architecture performs similar or better in the majority of the tested configurations.

Furthermore, we measured the average prediction latency and showed that \architecture leads to reduced latency even with a small percentage of remote calls avoided, due to the large difference in latency between local and remote inference. Overall, our results demonstrate that \architecture can be a cost-effective and accurate solution for systems relying on large-scale remote models, particularly on resource-constrained devices.

\section{Data Availability}
\label{sec:data_availabilty}

The full replication package for this paper is made available on Github, and archived on Zenodo:
\begin{itemize}
    \item \url{https://github.com/testingautomated-usi/bisupervised}
    \item \url{https://zenodo.org/record/7767383}
\end{itemize}

The replication package contains all scripts we used to calculate our results, as well as a Docker file allowing us to easily re-create the environment to run these scripts. The replication package also contains scripts to download and process third-party artifacts, such as models and datasets, automatically wherever possible. All our code and artifacts are permissively MIT licensed.
A corresponding TOSEM replicated computational results (RCR) report is currently under review.

\begin{acks}
This work was partially supported by the H2020 project PRECRIME,
funded under the ERC Advanced Grant 2017 Program (ERC Grant Agreement n. 787703).
\end{acks}

\bibliographystyle{ACM-Reference-Format}
\bibliography{main}


\begin{thebibliography}{73}


\ifx \showCODEN    \undefined \def \showCODEN     #1{\unskip}     \fi
\ifx \showDOI      \undefined \def \showDOI       #1{#1}\fi
\ifx \showISBNx    \undefined \def \showISBNx     #1{\unskip}     \fi
\ifx \showISBNxiii \undefined \def \showISBNxiii  #1{\unskip}     \fi
\ifx \showISSN     \undefined \def \showISSN      #1{\unskip}     \fi
\ifx \showLCCN     \undefined \def \showLCCN      #1{\unskip}     \fi
\ifx \shownote     \undefined \def \shownote      #1{#1}          \fi
\ifx \showarticletitle \undefined \def \showarticletitle #1{#1}   \fi
\ifx \showURL      \undefined \def \showURL       {\relax}        \fi
\providecommand\bibfield[2]{#2}
\providecommand\bibinfo[2]{#2}
\providecommand\natexlab[1]{#1}
\providecommand\showeprint[2][]{arXiv:#2}

\bibitem[Berend et~al\mbox{.}(2020)]%
        {Berend2020CatsAreNotFish}
\bibfield{author}{\bibinfo{person}{David Berend}, \bibinfo{person}{Xiaofei
  Xie}, \bibinfo{person}{Lei Ma}, \bibinfo{person}{Lingjun Zhou},
  \bibinfo{person}{Yang Liu}, \bibinfo{person}{Chi Xu}, {and}
  \bibinfo{person}{Jianjun Zhao}.} \bibinfo{year}{2020}\natexlab{}.
\newblock \showarticletitle{Cats Are Not Fish: Deep Learning Testing Calls for
  Out-Of-Distribution Awareness}. In \bibinfo{booktitle}{\emph{The 35th
  IEEE/ACM International Conference on Automated Software Engineering}}.
  \bibinfo{publisher}{Association for Computing Machinery},
  \bibinfo{address}{New York, NY, USA}.
\newblock


\bibitem[Brown et~al\mbox{.}(2020)]%
        {brown2020GPT3}
\bibfield{author}{\bibinfo{person}{Tom Brown}, \bibinfo{person}{Benjamin Mann},
  \bibinfo{person}{Nick Ryder}, \bibinfo{person}{Melanie Subbiah},
  \bibinfo{person}{Jared~D Kaplan}, \bibinfo{person}{Prafulla Dhariwal},
  \bibinfo{person}{Arvind Neelakantan}, \bibinfo{person}{Pranav Shyam},
  \bibinfo{person}{Girish Sastry}, \bibinfo{person}{Amanda Askell},
  {et~al\mbox{.}}} \bibinfo{year}{2020}\natexlab{}.
\newblock \showarticletitle{Language models are few-shot learners}.
\newblock \bibinfo{journal}{\emph{Advances in neural information processing
  systems}}  \bibinfo{volume}{33} (\bibinfo{year}{2020}),
  \bibinfo{pages}{1877--1901}.
\newblock


\bibitem[Carvalho et~al\mbox{.}(2020)]%
        {carvalho2020computationOffloadingUsingAI}
\bibfield{author}{\bibinfo{person}{Gon{\c{c}}alo Carvalho},
  \bibinfo{person}{Bruno Cabral}, \bibinfo{person}{Vasco Pereira}, {and}
  \bibinfo{person}{Jorge Bernardino}.} \bibinfo{year}{2020}\natexlab{}.
\newblock \showarticletitle{Computation offloading in Edge Computing
  environments using Artificial Intelligence techniques}.
\newblock \bibinfo{journal}{\emph{Engineering Applications of Artificial
  Intelligence}}  \bibinfo{volume}{95} (\bibinfo{year}{2020}),
  \bibinfo{pages}{103840}.
\newblock


\bibitem[Catak et~al\mbox{.}(2021)]%
        {Catak2021AutonomousDriving}
\bibfield{author}{\bibinfo{person}{Ferhat~Ozgur Catak}, \bibinfo{person}{Tao
  Yue}, {and} \bibinfo{person}{Shaukat Ali}.} \bibinfo{year}{2021}\natexlab{}.
\newblock \showarticletitle{Prediction Surface Uncertainty Quantification in
  Object Detection Models for Autonomous Driving}.
\newblock  (\bibinfo{year}{2021}).
\newblock
\showeprint[arXiv]{2107.04991v1}~[cs.CV]


\bibitem[Catak et~al\mbox{.}(2022)]%
        {Catak2022UncertaintyAware}
\bibfield{author}{\bibinfo{person}{Ferhat~Ozgur Catak}, \bibinfo{person}{Tao
  Yue}, {and} \bibinfo{person}{Shaukat Ali}.} \bibinfo{year}{2022}\natexlab{}.
\newblock \showarticletitle{Uncertainty-Aware Prediction Validator in Deep
  Learning Models for Cyber-Physical System Data}.
\newblock \bibinfo{journal}{\emph{ACM Trans. Softw. Eng. Methodol.}}
  \bibinfo{volume}{31}, \bibinfo{number}{4}, Article \bibinfo{articleno}{79}
  (\bibinfo{date}{jul} \bibinfo{year}{2022}), \bibinfo{numpages}{31}~pages.
\newblock
\showISSN{1049-331X}
\urldef\tempurl%
\url{https://doi.org/10.1145/3527451}
\showDOI{\tempurl}


\bibitem[Chen et~al\mbox{.}(2019)]%
        {Chen2019Antprophet}
\bibfield{author}{\bibinfo{person}{Cen Chen}, \bibinfo{person}{Xiaolu Zhang},
  \bibinfo{person}{Sheng Ju}, \bibinfo{person}{Chilin Fu},
  \bibinfo{person}{Caizhi Tang}, \bibinfo{person}{Jun Zhou}, {and}
  \bibinfo{person}{Xiaolong Li}.} \bibinfo{year}{2019}\natexlab{}.
\newblock \showarticletitle{AntProphet: an Intention Mining System behind
  Alipay's Intelligent Customer Service Bot.}. In
  \bibinfo{booktitle}{\emph{IJCAI}}, Vol.~\bibinfo{volume}{8}.
  \bibinfo{pages}{6497--6499}.
\newblock


\bibitem[Cire{\c{s}}an et~al\mbox{.}(2010)]%
        {cirecsan2010gpu}
\bibfield{author}{\bibinfo{person}{Dan~Claudiu Cire{\c{s}}an},
  \bibinfo{person}{Ueli Meier}, \bibinfo{person}{Luca~Maria Gambardella}, {and}
  \bibinfo{person}{J{\"u}rgen Schmidhuber}.} \bibinfo{year}{2010}\natexlab{}.
\newblock \showarticletitle{Deep, big, simple neural nets for handwritten digit
  recognition}.
\newblock \bibinfo{journal}{\emph{Neural computation}} \bibinfo{volume}{22},
  \bibinfo{number}{12} (\bibinfo{year}{2010}), \bibinfo{pages}{3207--3220}.
\newblock


\bibitem[David et~al\mbox{.}(2020)]%
        {david2020tflite}
\bibfield{author}{\bibinfo{person}{Robert David}, \bibinfo{person}{Jared Duke},
  \bibinfo{person}{Advait Jain}, \bibinfo{person}{Vijay~Janapa Reddi},
  \bibinfo{person}{Nat Jeffries}, \bibinfo{person}{Jian Li},
  \bibinfo{person}{Nick Kreeger}, \bibinfo{person}{Ian Nappier},
  \bibinfo{person}{Meghna Natraj}, \bibinfo{person}{Shlomi Regev},
  \bibinfo{person}{Rocky Rhodes}, \bibinfo{person}{Tiezhen Wang}, {and}
  \bibinfo{person}{Pete Warden}.} \bibinfo{year}{2020}\natexlab{}.
\newblock \bibinfo{title}{TensorFlow Lite Micro: Embedded Machine Learning on
  TinyML Systems}.
\newblock
\newblock
\urldef\tempurl%
\url{https://doi.org/10.48550/ARXIV.2010.08678}
\showDOI{\tempurl}


\bibitem[Devlin et~al\mbox{.}(2018)]%
        {devlin2018bert}
\bibfield{author}{\bibinfo{person}{Jacob Devlin}, \bibinfo{person}{Ming-Wei
  Chang}, \bibinfo{person}{Kenton Lee}, {and} \bibinfo{person}{Kristina
  Toutanova}.} \bibinfo{year}{2018}\natexlab{}.
\newblock \showarticletitle{Bert: Pre-training of deep bidirectional
  transformers for language understanding}.
\newblock \bibinfo{journal}{\emph{arXiv preprint arXiv:1810.04805}}
  (\bibinfo{year}{2018}).
\newblock


\bibitem[Dola et~al\mbox{.}(2021)]%
        {dola2021distribution}
\bibfield{author}{\bibinfo{person}{Swaroopa Dola}, \bibinfo{person}{Matthew~B
  Dwyer}, {and} \bibinfo{person}{Mary~Lou Soffa}.}
  \bibinfo{year}{2021}\natexlab{}.
\newblock \showarticletitle{Distribution-aware testing of neural networks using
  generative models}. In \bibinfo{booktitle}{\emph{2021 IEEE/ACM 43rd
  International Conference on Software Engineering (ICSE)}}. IEEE,
  \bibinfo{pages}{226--237}.
\newblock


\bibitem[Eshed(2020)]%
        {Eshed2020ImagenetNoveltyDetection}
\bibfield{author}{\bibinfo{person}{Noam Eshed}.}
  \bibinfo{year}{2020}\natexlab{}.
\newblock \emph{\bibinfo{title}{Novelty detection and analysis in convolutional
  neural networks}}.
\newblock \bibinfo{thesistype}{Master's\ thesis}. \bibinfo{school}{Cornell
  University}.
\newblock


\bibitem[Fedus et~al\mbox{.}(2021)]%
        {fedus2021switch}
\bibfield{author}{\bibinfo{person}{William Fedus}, \bibinfo{person}{Barret
  Zoph}, {and} \bibinfo{person}{Noam Shazeer}.}
  \bibinfo{year}{2021}\natexlab{}.
\newblock \bibinfo{title}{Switch transformers: Scaling to trillion parameter
  models with simple and efficient sparsity}.
\newblock
\newblock


\bibitem[Feng et~al\mbox{.}(2020)]%
        {feng2020deepgini}
\bibfield{author}{\bibinfo{person}{Yang Feng}, \bibinfo{person}{Qingkai Shi},
  \bibinfo{person}{Xinyu Gao}, \bibinfo{person}{Jun Wan},
  \bibinfo{person}{Chunrong Fang}, {and} \bibinfo{person}{Zhenyu Chen}.}
  \bibinfo{year}{2020}\natexlab{}.
\newblock \showarticletitle{Deepgini: prioritizing massive tests to enhance the
  robustness of deep neural networks}. In \bibinfo{booktitle}{\emph{Proceedings
  of the 29th ACM SIGSOFT International Symposium on Software Testing and
  Analysis}}. \bibinfo{pages}{177--188}.
\newblock


\bibitem[Gal and Ghahramani(2016)]%
        {gal2016dropout}
\bibfield{author}{\bibinfo{person}{Yarin Gal} {and} \bibinfo{person}{Zoubin
  Ghahramani}.} \bibinfo{year}{2016}\natexlab{}.
\newblock \showarticletitle{Dropout as a bayesian approximation: Representing
  model uncertainty in deep learning}. In
  \bibinfo{booktitle}{\emph{international conference on machine learning}}.
  PMLR, \bibinfo{pages}{1050--1059}.
\newblock


\bibitem[Hendrycks et~al\mbox{.}(2021)]%
        {hendrycks2021unsolved}
\bibfield{author}{\bibinfo{person}{Dan Hendrycks}, \bibinfo{person}{Nicholas
  Carlini}, \bibinfo{person}{John Schulman}, {and} \bibinfo{person}{Jacob
  Steinhardt}.} \bibinfo{year}{2021}\natexlab{}.
\newblock \showarticletitle{Unsolved problems in ml safety}.
\newblock \bibinfo{journal}{\emph{arXiv preprint arXiv:2109.13916}}
  (\bibinfo{year}{2021}).
\newblock


\bibitem[Hendrycks and Gimpel(2016)]%
        {Hendrycks2016Softmax}
\bibfield{author}{\bibinfo{person}{Dan Hendrycks} {and} \bibinfo{person}{Kevin
  Gimpel}.} \bibinfo{year}{2016}\natexlab{}.
\newblock \showarticletitle{A Baseline for Detecting Misclassified and
  Out-of-Distribution Examples in Neural Networks}.
\newblock  (\bibinfo{year}{2016}).
\newblock
\showeprint[arXiv]{1610.02136v3}~[cs.NE]


\bibitem[Hendrycks et~al\mbox{.}(2019)]%
        {Hendrycks2019OutlierExposure}
\bibfield{author}{\bibinfo{person}{Dan Hendrycks}, \bibinfo{person}{Mantas
  Mazeika}, {and} \bibinfo{person}{Thomas Dietterich}.}
  \bibinfo{year}{2019}\natexlab{}.
\newblock \showarticletitle{Deep Anomaly Detection with Outlier Exposure}.
\newblock  (\bibinfo{year}{2019}).
\newblock
\showeprint[arXiv]{1812.04606v3}~[cs.LG]


\bibitem[Howard et~al\mbox{.}(2019)]%
        {howard2019mobilenetv3}
\bibfield{author}{\bibinfo{person}{Andrew Howard}, \bibinfo{person}{Mark
  Sandler}, \bibinfo{person}{Grace Chu}, \bibinfo{person}{Liang-Chieh Chen},
  \bibinfo{person}{Bo Chen}, \bibinfo{person}{Mingxing Tan},
  \bibinfo{person}{Weijun Wang}, \bibinfo{person}{Yukun Zhu},
  \bibinfo{person}{Ruoming Pang}, \bibinfo{person}{Vijay Vasudevan},
  {et~al\mbox{.}}} \bibinfo{year}{2019}\natexlab{}.
\newblock \showarticletitle{Searching for mobilenetv3}. In
  \bibinfo{booktitle}{\emph{Proceedings of the IEEE/CVF international
  conference on computer vision}}. \bibinfo{pages}{1314--1324}.
\newblock


\bibitem[Howard et~al\mbox{.}(2017)]%
        {howard2017mobilenets}
\bibfield{author}{\bibinfo{person}{Andrew~G Howard}, \bibinfo{person}{Menglong
  Zhu}, \bibinfo{person}{Bo Chen}, \bibinfo{person}{Dmitry Kalenichenko},
  \bibinfo{person}{Weijun Wang}, \bibinfo{person}{Tobias Weyand},
  \bibinfo{person}{Marco Andreetto}, {and} \bibinfo{person}{Hartwig Adam}.}
  \bibinfo{year}{2017}\natexlab{}.
\newblock \showarticletitle{Mobilenets: Efficient convolutional neural networks
  for mobile vision applications}.
\newblock \bibinfo{journal}{\emph{arXiv preprint arXiv:1704.04861}}
  (\bibinfo{year}{2017}).
\newblock


\bibitem[Ibrahim and Abdulazeez(2021)]%
        {ibrahim2021mlDiagnosis}
\bibfield{author}{\bibinfo{person}{Ibrahim Ibrahim} {and}
  \bibinfo{person}{Adnan Abdulazeez}.} \bibinfo{year}{2021}\natexlab{}.
\newblock \showarticletitle{The role of machine learning algorithms for
  diagnosing diseases}.
\newblock \bibinfo{journal}{\emph{Journal of Applied Science and Technology
  Trends}} \bibinfo{volume}{2}, \bibinfo{number}{01} (\bibinfo{year}{2021}),
  \bibinfo{pages}{10--19}.
\newblock


\bibitem[Izadi(2022)]%
        {izadi2022catiss}
\bibfield{author}{\bibinfo{person}{Maliheh Izadi}.}
  \bibinfo{year}{2022}\natexlab{}.
\newblock \showarticletitle{CatIss: An Intelligent Tool for Categorizing Issues
  Reports using Transformers}.
\newblock \bibinfo{journal}{\emph{arXiv preprint arXiv:2203.17196}}
  (\bibinfo{year}{2022}).
\newblock


\bibitem[Jiao et~al\mbox{.}(2019)]%
        {jiao2019tinybert}
\bibfield{author}{\bibinfo{person}{Xiaoqi Jiao}, \bibinfo{person}{Yichun Yin},
  \bibinfo{person}{Lifeng Shang}, \bibinfo{person}{Xin Jiang},
  \bibinfo{person}{Xiao Chen}, \bibinfo{person}{Linlin Li},
  \bibinfo{person}{Fang Wang}, {and} \bibinfo{person}{Qun Liu}.}
  \bibinfo{year}{2019}\natexlab{}.
\newblock \showarticletitle{Tinybert: Distilling bert for natural language
  understanding}.
\newblock \bibinfo{journal}{\emph{arXiv preprint arXiv:1909.10351}}
  (\bibinfo{year}{2019}).
\newblock


\bibitem[Jospin et~al\mbox{.}(2022)]%
        {jospin2022bayesian}
\bibfield{author}{\bibinfo{person}{Laurent~Valentin Jospin},
  \bibinfo{person}{Hamid Laga}, \bibinfo{person}{Farid Boussaid},
  \bibinfo{person}{Wray Buntine}, {and} \bibinfo{person}{Mohammed Bennamoun}.}
  \bibinfo{year}{2022}\natexlab{}.
\newblock \showarticletitle{Hands-on Bayesian neural networks—A tutorial for
  deep learning users}.
\newblock \bibinfo{journal}{\emph{IEEE Computational Intelligence Magazine}}
  \bibinfo{volume}{17}, \bibinfo{number}{2} (\bibinfo{year}{2022}),
  \bibinfo{pages}{29--48}.
\newblock


\bibitem[Kallis et~al\mbox{.}(2022)]%
        {kallis2022nlbse}
\bibfield{author}{\bibinfo{person}{Rafael Kallis}, \bibinfo{person}{Oscar
  Chaparro}, \bibinfo{person}{Andrea Di~Sorbo}, {and}
  \bibinfo{person}{Sebastiano Panichella}.} \bibinfo{year}{2022}\natexlab{}.
\newblock \showarticletitle{Nlbse’22 tool competition}. In
  \bibinfo{booktitle}{\emph{2022 IEEE/ACM 1st International Workshop on Natural
  Language-Based Software Engineering (NLBSE)}}. IEEE, \bibinfo{pages}{25--28}.
\newblock


\bibitem[Kallis et~al\mbox{.}(2019)]%
        {kallis2019ticketTagger}
\bibfield{author}{\bibinfo{person}{Rafael Kallis}, \bibinfo{person}{Andrea
  Di~Sorbo}, \bibinfo{person}{Gerardo Canfora}, {and}
  \bibinfo{person}{Sebastiano Panichella}.} \bibinfo{year}{2019}\natexlab{}.
\newblock \showarticletitle{Ticket tagger: Machine learning driven issue
  classification}. In \bibinfo{booktitle}{\emph{2019 IEEE International
  Conference on Software Maintenance and Evolution (ICSME)}}. IEEE,
  \bibinfo{pages}{406--409}.
\newblock


\bibitem[Karpathy(2022)]%
        {karpathy2022Lecun}
\bibfield{author}{\bibinfo{person}{Andrej Karpathy}.}
  \bibinfo{year}{2022}\natexlab{}.
\newblock \bibinfo{booktitle}{\emph{Deep Neural Nets: 33 years ago and 33 years
  from now}}.
\newblock
\urldef\tempurl%
\url{http://karpathy.github.io/2022/03/14/lecun1989/}
\showURL{%
\tempurl}


\bibitem[Kendall and Gal(2017)]%
        {kendall2017uncertainties}
\bibfield{author}{\bibinfo{person}{Alex Kendall} {and} \bibinfo{person}{Yarin
  Gal}.} \bibinfo{year}{2017}\natexlab{}.
\newblock \showarticletitle{What uncertainties do we need in bayesian deep
  learning for computer vision?}
\newblock \bibinfo{journal}{\emph{Advances in neural information processing
  systems}}  \bibinfo{volume}{30} (\bibinfo{year}{2017}).
\newblock


\bibitem[Kim et~al\mbox{.}(2019)]%
        {Kim2018SurpriseAdequacy}
\bibfield{author}{\bibinfo{person}{Jinhan Kim}, \bibinfo{person}{Robert Feldt},
  {and} \bibinfo{person}{Shin Yoo}.} \bibinfo{year}{2019}\natexlab{}.
\newblock \showarticletitle{Guiding deep learning system testing using surprise
  adequacy}. In \bibinfo{booktitle}{\emph{2019 IEEE/ACM 41st International
  Conference on Software Engineering (ICSE)}}. \bibinfo{pages}{1039--1049}.
\newblock


\bibitem[Kim et~al\mbox{.}(2020)]%
        {Kim2020ReducingLabellingCost}
\bibfield{author}{\bibinfo{person}{Jinhan Kim}, \bibinfo{person}{Jeongil Ju},
  \bibinfo{person}{Robert Feldt}, {and} \bibinfo{person}{Shin Yoo}.}
  \bibinfo{year}{2020}\natexlab{}.
\newblock \showarticletitle{Reducing dnn labelling cost using surprise
  adequacy: An industrial case study for autonomous driving}. In
  \bibinfo{booktitle}{\emph{Proceedings of the 28th ACM Joint Meeting on
  European Software Engineering Conference and Symposium on the Foundations of
  Software Engineering}}. \bibinfo{pages}{1466--1476}.
\newblock


\bibitem[Kim and Yoo(2020)]%
        {Kim2020EvaluatingSAforQA}
\bibfield{author}{\bibinfo{person}{Seah Kim} {and} \bibinfo{person}{Shin Yoo}.}
  \bibinfo{year}{2020}\natexlab{}.
\newblock \showarticletitle{Evaluating surprise adequacy for question
  answering}. In \bibinfo{booktitle}{\emph{Proceedings of the IEEE/ACM 42nd
  International Conference on Software Engineering Workshops}}.
  \bibinfo{pages}{197--202}.
\newblock


\bibitem[Kim and Yoo(2021)]%
        {Kim2021MultiModalSA}
\bibfield{author}{\bibinfo{person}{S. Kim} {and} \bibinfo{person}{S. Yoo}.}
  \bibinfo{year}{2021}\natexlab{}.
\newblock \showarticletitle{Multimodal Surprise Adequacy Analysis of Inputs for
  Natural Language Processing DNN Models}. In \bibinfo{booktitle}{\emph{2021
  IEEE/ACM International Conference on Automation of Software Test (AST)}}.
  \bibinfo{publisher}{IEEE Computer Society}, \bibinfo{address}{Los Alamitos,
  CA, USA}, \bibinfo{pages}{80--89}.
\newblock
\urldef\tempurl%
\url{https://doi.org/10.1109/AST52587.2021.00017}
\showDOI{\tempurl}


\bibitem[Krizhevsky et~al\mbox{.}(2012)]%
        {Krizhevsky2012Alexnet}
\bibfield{author}{\bibinfo{person}{Alex Krizhevsky}, \bibinfo{person}{Ilya
  Sutskever}, {and} \bibinfo{person}{Geoffrey~E Hinton}.}
  \bibinfo{year}{2012}\natexlab{}.
\newblock \showarticletitle{ImageNet Classification with Deep Convolutional
  Neural Networks}. In \bibinfo{booktitle}{\emph{Advances in Neural Information
  Processing Systems}}, \bibfield{editor}{\bibinfo{person}{F.~Pereira},
  \bibinfo{person}{C.J. Burges}, \bibinfo{person}{L.~Bottou}, {and}
  \bibinfo{person}{K.Q. Weinberger}} (Eds.), Vol.~\bibinfo{volume}{25}.
  \bibinfo{publisher}{Curran Associates, Inc.}
\newblock
\urldef\tempurl%
\url{https://proceedings.neurips.cc/paper/2012/file/c399862d3b9d6b76c8436e924a68c45b-Paper.pdf}
\showURL{%
\tempurl}


\bibitem[Lakshminarayanan et~al\mbox{.}(2017)]%
        {Lakshminarayanan2017Ensembles}
\bibfield{author}{\bibinfo{person}{Balaji Lakshminarayanan},
  \bibinfo{person}{Alexander Pritzel}, {and} \bibinfo{person}{Charles
  Blundell}.} \bibinfo{year}{2017}\natexlab{}.
\newblock \showarticletitle{Simple and scalable predictive uncertainty
  estimation using deep ensembles}. In \bibinfo{booktitle}{\emph{Advances in
  neural information processing systems}}. \bibinfo{pages}{6402--6413}.
\newblock


\bibitem[LeCun et~al\mbox{.}(1989)]%
        {lecun1989backpropagation}
\bibfield{author}{\bibinfo{person}{Yann LeCun}, \bibinfo{person}{Bernhard
  Boser}, \bibinfo{person}{John~S Denker}, \bibinfo{person}{Donnie Henderson},
  \bibinfo{person}{Richard~E Howard}, \bibinfo{person}{Wayne Hubbard}, {and}
  \bibinfo{person}{Lawrence~D Jackel}.} \bibinfo{year}{1989}\natexlab{}.
\newblock \showarticletitle{Backpropagation applied to handwritten zip code
  recognition}.
\newblock \bibinfo{journal}{\emph{Neural computation}} \bibinfo{volume}{1},
  \bibinfo{number}{4} (\bibinfo{year}{1989}), \bibinfo{pages}{541--551}.
\newblock


\bibitem[Liu et~al\mbox{.}(2020)]%
        {liu2020autocompress}
\bibfield{author}{\bibinfo{person}{Ning Liu}, \bibinfo{person}{Xiaolong Ma},
  \bibinfo{person}{Zhiyuan Xu}, \bibinfo{person}{Yanzhi Wang},
  \bibinfo{person}{Jian Tang}, {and} \bibinfo{person}{Jieping Ye}.}
  \bibinfo{year}{2020}\natexlab{}.
\newblock \showarticletitle{Autocompress: An automatic dnn structured pruning
  framework for ultra-high compression rates}. In
  \bibinfo{booktitle}{\emph{Proceedings of the AAAI Conference on Artificial
  Intelligence}}, Vol.~\bibinfo{volume}{34}. \bibinfo{pages}{4876--4883}.
\newblock


\bibitem[Liu et~al\mbox{.}(2019)]%
        {Liu2019Roberta}
\bibfield{author}{\bibinfo{person}{Yinhan Liu}, \bibinfo{person}{Myle Ott},
  \bibinfo{person}{Naman Goyal}, \bibinfo{person}{Jingfei Du},
  \bibinfo{person}{Mandar Joshi}, \bibinfo{person}{Danqi Chen},
  \bibinfo{person}{Omer Levy}, \bibinfo{person}{Mike Lewis},
  \bibinfo{person}{Luke Zettlemoyer}, {and} \bibinfo{person}{Veselin
  Stoyanov}.} \bibinfo{year}{2019}\natexlab{}.
\newblock \showarticletitle{RoBERTa: {A} Robustly Optimized {BERT} Pretraining
  Approach}.
\newblock \bibinfo{journal}{\emph{CoRR}}  \bibinfo{volume}{abs/1907.11692}
  (\bibinfo{year}{2019}).
\newblock
\showeprint[arxiv]{1907.11692}
\urldef\tempurl%
\url{http://arxiv.org/abs/1907.11692}
\showURL{%
\tempurl}


\bibitem[Lu et~al\mbox{.}(2019)]%
        {Lu2019LearntAssertions}
\bibfield{author}{\bibinfo{person}{Haochuan Lu}, \bibinfo{person}{Huanlin Xu},
  \bibinfo{person}{Nana Liu}, \bibinfo{person}{Yangfan Zhou}, {and}
  \bibinfo{person}{Xin Wang}.} \bibinfo{year}{2019}\natexlab{}.
\newblock \showarticletitle{Data Sanity Check for Deep Learning Systems via
  Learnt Assertions}.
\newblock  (\bibinfo{year}{2019}).
\newblock
\showeprint[arXiv]{1909.03835v3}~[cs.LG]


\bibitem[Maas et~al\mbox{.}(2011)]%
        {Maas2011imdb}
\bibfield{author}{\bibinfo{person}{Andrew~L. Maas}, \bibinfo{person}{Raymond~E.
  Daly}, \bibinfo{person}{Peter~T. Pham}, \bibinfo{person}{Dan Huang},
  \bibinfo{person}{Andrew~Y. Ng}, {and} \bibinfo{person}{Christopher Potts}.}
  \bibinfo{year}{2011}\natexlab{}.
\newblock \showarticletitle{Learning Word Vectors for Sentiment Analysis}. In
  \bibinfo{booktitle}{\emph{Proceedings of the 49th Annual Meeting of the
  Association for Computational Linguistics: Human Language Technologies}}.
  \bibinfo{publisher}{Association for Computational Linguistics},
  \bibinfo{address}{Portland, Oregon, USA}, \bibinfo{pages}{142--150}.
\newblock
\urldef\tempurl%
\url{http://www.aclweb.org/anthology/P11-1015}
\showURL{%
\tempurl}


\bibitem[Mattina(2020)]%
        {mattina2020Cost}
\bibfield{author}{\bibinfo{person}{Matthew Mattina}.}
  \bibinfo{year}{2020}\natexlab{}.
\newblock \bibinfo{booktitle}{\emph{Reducing the Cost of Neural Network
  Inference with Residue Number Systems}}.
\newblock
\urldef\tempurl%
\url{https://community.arm.com/arm-research/b/articles/posts/reducing-the-cost-of-neural-network-inference-with-residue-number-systems}
\showURL{%
\tempurl}


\bibitem[Michelmore et~al\mbox{.}(2018)]%
        {Michelmore2018Autonomous}
\bibfield{author}{\bibinfo{person}{Rhiannon Michelmore}, \bibinfo{person}{Marta
  Kwiatkowska}, {and} \bibinfo{person}{Yarin Gal}.}
  \bibinfo{year}{2018}\natexlab{}.
\newblock \showarticletitle{Evaluating Uncertainty Quantification in End-to-End
  Autonomous Driving Control}.
\newblock \bibinfo{journal}{\emph{CoRR}} (\bibinfo{year}{2018}).
\newblock
\showeprint[arXiv]{1811.06817v1}~[cs.LG]


\bibitem[Ovadia et~al\mbox{.}(2019)]%
        {Ovadia2019}
\bibfield{author}{\bibinfo{person}{Yaniv Ovadia}, \bibinfo{person}{Emily
  Fertig}, \bibinfo{person}{Jie Ren}, \bibinfo{person}{Zachary Nado},
  \bibinfo{person}{D. Sculley}, \bibinfo{person}{Sebastian Nowozin},
  \bibinfo{person}{Joshua Dillon}, \bibinfo{person}{Balaji Lakshminarayanan},
  {and} \bibinfo{person}{Jasper Snoek}.} \bibinfo{year}{2019}\natexlab{}.
\newblock \showarticletitle{Can you trust your models uncertainty? Evaluating
  predictive uncertainty under dataset shift}.
\newblock \bibinfo{journal}{\emph{Advances in Neural Information Processing
  Systems}} (\bibinfo{year}{2019}), \bibinfo{pages}{13991--14002}.
\newblock


\bibitem[Peng and Budhkar(2021)]%
        {GP3ParamEstimate}
\bibfield{author}{\bibinfo{person}{Zilun Peng} {and} \bibinfo{person}{Akshay
  Budhkar}.} \bibinfo{year}{2021}\natexlab{}.
\newblock \bibinfo{booktitle}{\emph{GPT-Neo vs. GPT-3: Are Commercialized NLP
  Models Really That Much Better?}}
\newblock
\urldef\tempurl%
\url{https://medium.com/georgian-impact-blog/gpt-neo-vs-gpt-3-are-commercialized-nlp-models-really-that-much-better-f4c73ffce10b}
\showURL{%
\tempurl}


\bibitem[Radford et~al\mbox{.}(2019)]%
        {radford2019GPT2}
\bibfield{author}{\bibinfo{person}{Alec Radford}, \bibinfo{person}{Jeffrey Wu},
  \bibinfo{person}{Rewon Child}, \bibinfo{person}{David Luan},
  \bibinfo{person}{Dario Amodei}, \bibinfo{person}{Ilya Sutskever},
  {et~al\mbox{.}}} \bibinfo{year}{2019}\natexlab{}.
\newblock \showarticletitle{Language models are unsupervised multitask
  learners}.
\newblock \bibinfo{journal}{\emph{OpenAI blog}} \bibinfo{volume}{1},
  \bibinfo{number}{8} (\bibinfo{year}{2019}), \bibinfo{pages}{9}.
\newblock


\bibitem[Rajpurkar et~al\mbox{.}(2018)]%
        {rajpurkar2018know}
\bibfield{author}{\bibinfo{person}{Pranav Rajpurkar}, \bibinfo{person}{Robin
  Jia}, {and} \bibinfo{person}{Percy Liang}.} \bibinfo{year}{2018}\natexlab{}.
\newblock \showarticletitle{Know what you don't know: Unanswerable questions
  for SQuAD}.
\newblock \bibinfo{journal}{\emph{arXiv preprint arXiv:1806.03822}}
  (\bibinfo{year}{2018}).
\newblock


\bibitem[Riccio et~al\mbox{.}(2020)]%
        {Riccio2020}
\bibfield{author}{\bibinfo{person}{Vincenzo Riccio}, \bibinfo{person}{Gunel
  Jahangiroba}, \bibinfo{person}{Andrea Stocco}, \bibinfo{person}{Nargiz
  Humbatova}, \bibinfo{person}{Michael Weiss}, {and} \bibinfo{person}{Paolo
  Tonella}.} \bibinfo{year}{2020}\natexlab{}.
\newblock \showarticletitle{Testing machine learning based systems: a
  systematic mapping}.
\newblock \bibinfo{journal}{\emph{Empirical Software Engineering}}
  (\bibinfo{year}{2020}).
\newblock


\bibitem[Riccio and Tonella(2023)]%
        {riccio2023validators}
\bibfield{author}{\bibinfo{person}{Vincenzo Riccio} {and}
  \bibinfo{person}{Paolo Tonella}.} \bibinfo{year}{2023}\natexlab{}.
\newblock \showarticletitle{When and Why Test Generators for Deep Learning
  Produce Invalid Inputs: an Empirical Study}. In
  \bibinfo{booktitle}{\emph{2023 IEEE/ACM 45th International Conference on
  Software Engineering (ICSE)}}.
\newblock
\newblock
\shownote{forthcoming}.


\bibitem[Russakovsky et~al\mbox{.}(2015)]%
        {russakovsky2015imagenet}
\bibfield{author}{\bibinfo{person}{Olga Russakovsky}, \bibinfo{person}{Jia
  Deng}, \bibinfo{person}{Hao Su}, \bibinfo{person}{Jonathan Krause},
  \bibinfo{person}{Sanjeev Satheesh}, \bibinfo{person}{Sean Ma},
  \bibinfo{person}{Zhiheng Huang}, \bibinfo{person}{Andrej Karpathy},
  \bibinfo{person}{Aditya Khosla}, \bibinfo{person}{Michael Bernstein},
  {et~al\mbox{.}}} \bibinfo{year}{2015}\natexlab{}.
\newblock \showarticletitle{Imagenet large scale visual recognition challenge}.
\newblock \bibinfo{journal}{\emph{International journal of computer vision}}
  \bibinfo{volume}{115}, \bibinfo{number}{3} (\bibinfo{year}{2015}),
  \bibinfo{pages}{211--252}.
\newblock


\bibitem[Saeik et~al\mbox{.}(2021)]%
        {saeik2021taskOffloadingSurvey}
\bibfield{author}{\bibinfo{person}{Firdose Saeik}, \bibinfo{person}{Marios
  Avgeris}, \bibinfo{person}{Dimitrios Spatharakis}, \bibinfo{person}{Nina
  Santi}, \bibinfo{person}{Dimitrios Dechouniotis}, \bibinfo{person}{John
  Violos}, \bibinfo{person}{Aris Leivadeas}, \bibinfo{person}{Nikolaos
  Athanasopoulos}, \bibinfo{person}{Nathalie Mitton}, {and}
  \bibinfo{person}{Symeon Papavassiliou}.} \bibinfo{year}{2021}\natexlab{}.
\newblock \showarticletitle{Task offloading in Edge and Cloud Computing: A
  survey on mathematical, artificial intelligence and control theory
  solutions}.
\newblock \bibinfo{journal}{\emph{Computer Networks}}  \bibinfo{volume}{195}
  (\bibinfo{year}{2021}), \bibinfo{pages}{108177}.
\newblock


\bibitem[Sandler et~al\mbox{.}(2018)]%
        {sandler2018mobilenetv2}
\bibfield{author}{\bibinfo{person}{Mark Sandler}, \bibinfo{person}{Andrew
  Howard}, \bibinfo{person}{Menglong Zhu}, \bibinfo{person}{Andrey Zhmoginov},
  {and} \bibinfo{person}{Liang-Chieh Chen}.} \bibinfo{year}{2018}\natexlab{}.
\newblock \showarticletitle{Mobilenetv2: Inverted residuals and linear
  bottlenecks}. In \bibinfo{booktitle}{\emph{Proceedings of the IEEE conference
  on computer vision and pattern recognition}}. \bibinfo{pages}{4510--4520}.
\newblock


\bibitem[Smith et~al\mbox{.}(2022)]%
        {smith2022megatron}
\bibfield{author}{\bibinfo{person}{Shaden Smith}, \bibinfo{person}{Mostofa
  Patwary}, \bibinfo{person}{Brandon Norick}, \bibinfo{person}{Patrick
  LeGresley}, \bibinfo{person}{Samyam Rajbhandari}, \bibinfo{person}{Jared
  Casper}, \bibinfo{person}{Zhun Liu}, \bibinfo{person}{Shrimai Prabhumoye},
  \bibinfo{person}{George Zerveas}, \bibinfo{person}{Vijay Korthikanti},
  {et~al\mbox{.}}} \bibinfo{year}{2022}\natexlab{}.
\newblock \showarticletitle{Using deepspeed and megatron to train
  megatron-turing nlg 530b, a large-scale generative language model}.
\newblock \bibinfo{journal}{\emph{arXiv preprint arXiv:2201.11990}}
  (\bibinfo{year}{2022}).
\newblock


\bibitem[Srivastava et~al\mbox{.}(2014)]%
        {srivastava2014dropout}
\bibfield{author}{\bibinfo{person}{Nitish Srivastava},
  \bibinfo{person}{Geoffrey Hinton}, \bibinfo{person}{Alex Krizhevsky},
  \bibinfo{person}{Ilya Sutskever}, {and} \bibinfo{person}{Ruslan
  Salakhutdinov}.} \bibinfo{year}{2014}\natexlab{}.
\newblock \showarticletitle{Dropout: A Simple Way to Prevent Neural Networks
  from Overfitting}.
\newblock \bibinfo{journal}{\emph{Journal of Machine Learning Research}}
  \bibinfo{volume}{15}, \bibinfo{number}{56} (\bibinfo{year}{2014}),
  \bibinfo{pages}{1929--1958}.
\newblock
\urldef\tempurl%
\url{http://jmlr.org/papers/v15/srivastava14a.html}
\showURL{%
\tempurl}


\bibitem[Stocco and Tonella(2020)]%
        {stocco2020towardsContinuously}
\bibfield{author}{\bibinfo{person}{Andrea Stocco} {and} \bibinfo{person}{Paolo
  Tonella}.} \bibinfo{year}{2020}\natexlab{}.
\newblock \showarticletitle{Towards anomaly detectors that learn continuously}.
  In \bibinfo{booktitle}{\emph{2020 IEEE International Symposium on Software
  Reliability Engineering Workshops (ISSREW)}}. IEEE,
  \bibinfo{pages}{201--208}.
\newblock


\bibitem[Stocco and Tonella(2022)]%
        {stocco2021confidenceDrivenAE}
\bibfield{author}{\bibinfo{person}{Andrea Stocco} {and} \bibinfo{person}{Paolo
  Tonella}.} \bibinfo{year}{2022}\natexlab{}.
\newblock \showarticletitle{Confidence-driven weighted retraining for
  predicting safety-critical failures in autonomous driving systems}.
\newblock \bibinfo{journal}{\emph{Journal of Software: Evolution and Process}}
  \bibinfo{volume}{34}, \bibinfo{number}{10} (\bibinfo{year}{2022}),
  \bibinfo{pages}{e2386}.
\newblock


\bibitem[Stocco et~al\mbox{.}(2020a)]%
        {Stocco2020}
\bibfield{author}{\bibinfo{person}{Andrea Stocco}, \bibinfo{person}{Michael
  Weiss}, \bibinfo{person}{Marco Calzana}, {and} \bibinfo{person}{Paolo
  Tonella}.} \bibinfo{year}{2020}\natexlab{a}.
\newblock \showarticletitle{Misbehaviour Prediction for Autonomous Driving
  Systems}. In \bibinfo{booktitle}{\emph{Proceedings of 42nd International
  Conference on Software Engineering}}. \bibinfo{publisher}{ACM},
  \bibinfo{pages}{12 pages}.
\newblock


\bibitem[Stocco et~al\mbox{.}(2020b)]%
        {Stocco2020Poster}
\bibfield{author}{\bibinfo{person}{Andrea Stocco}, \bibinfo{person}{Michael
  Weiss}, \bibinfo{person}{Marco Calzana}, {and} \bibinfo{person}{Paolo
  Tonella}.} \bibinfo{year}{2020}\natexlab{b}.
\newblock \showarticletitle{Predicting Safety-Critical Misbehaviours in
  Autonomous Driving Systems using Autoencoders}. In
  \bibinfo{booktitle}{\emph{Proceedings of 42nd International Conference on
  Software Engineering - Poster Track}}.
\newblock


\bibitem[Tan and Le(2019)]%
        {tan2019efficientnet}
\bibfield{author}{\bibinfo{person}{Mingxing Tan} {and} \bibinfo{person}{Quoc
  Le}.} \bibinfo{year}{2019}\natexlab{}.
\newblock \showarticletitle{Efficientnet: Rethinking model scaling for
  convolutional neural networks}. In \bibinfo{booktitle}{\emph{International
  conference on machine learning}}. PMLR, \bibinfo{pages}{6105--6114}.
\newblock


\bibitem[Tan and Le(2021)]%
        {tan2021efficientnetv2}
\bibfield{author}{\bibinfo{person}{Mingxing Tan} {and} \bibinfo{person}{Quoc
  Le}.} \bibinfo{year}{2021}\natexlab{}.
\newblock \showarticletitle{Efficientnetv2: Smaller models and faster
  training}. In \bibinfo{booktitle}{\emph{International Conference on Machine
  Learning}}. PMLR, \bibinfo{pages}{10096--10106}.
\newblock


\bibitem[Turc et~al\mbox{.}(2019)]%
        {turc2019bertTiny}
\bibfield{author}{\bibinfo{person}{Iulia Turc}, \bibinfo{person}{Ming-Wei
  Chang}, \bibinfo{person}{Kenton Lee}, {and} \bibinfo{person}{Kristina
  Toutanova}.} \bibinfo{year}{2019}\natexlab{}.
\newblock \showarticletitle{Well-read students learn better: On the importance
  of pre-training compact models}.
\newblock \bibinfo{journal}{\emph{arXiv preprint arXiv:1908.08962}}
  (\bibinfo{year}{2019}).
\newblock


\bibitem[Wang et~al\mbox{.}(2020)]%
        {Wang2020Dissector}
\bibfield{author}{\bibinfo{person}{Huiyan Wang}, \bibinfo{person}{Jingwei Xu},
  \bibinfo{person}{Chang Xu}, \bibinfo{person}{Xiaoxing Ma}, {and}
  \bibinfo{person}{Jian Lu}.} \bibinfo{year}{2020}\natexlab{}.
\newblock \showarticletitle{DISSECTOR: Input Validation for Deep Learning
  Applications by Crossing-layer Dissection}. In
  \bibinfo{booktitle}{\emph{Proceedings of 42nd International Conference on
  Software Engineering}}. \bibinfo{publisher}{ACM}.
\newblock


\bibitem[Weinreich and Pl{\"o}sch(2003)]%
        {weinreich2003remote}
\bibfield{author}{\bibinfo{person}{Rainer Weinreich} {and}
  \bibinfo{person}{Reinhold Pl{\"o}sch}.} \bibinfo{year}{2003}\natexlab{}.
\newblock \showarticletitle{Remote Configuration of Agent-Based Component
  Systems}.
\newblock \bibinfo{journal}{\emph{Journal of Object Technology}}
  \bibinfo{volume}{2}, \bibinfo{number}{6} (\bibinfo{year}{2003}),
  \bibinfo{pages}{67--84}.
\newblock


\bibitem[Weiss(2022)]%
        {Weiss2022CheapET3}
\bibfield{author}{\bibinfo{person}{Michael Weiss}.}
  \bibinfo{year}{2022}\natexlab{}.
\newblock \showarticletitle{CheapET-3: Cost-Efficient Use of Remote DNN
  Models}. In \bibinfo{booktitle}{\emph{Proceedings of the 30th ACM Joint
  European Software Engineering Conference and Symposium on the Foundations of
  Software Engineering (ESEC/FSE '22)}}.
\newblock


\bibitem[Weiss et~al\mbox{.}(2021)]%
        {Weiss2021-SA}
\bibfield{author}{\bibinfo{person}{Michael Weiss}, \bibinfo{person}{Rwiddhi
  Chakraborty}, {and} \bibinfo{person}{Paolo Tonella}.}
  \bibinfo{year}{2021}\natexlab{}.
\newblock \showarticletitle{A Review and Refinement of Surprise Adequacy}. In
  \bibinfo{booktitle}{\emph{2021 IEEE/ACM Third International Workshop on Deep
  Learning for Testing and Testing for Deep Learning (DeepTest)}}. IEEE,
  \bibinfo{pages}{17--24}.
\newblock


\bibitem[Weiss et~al\mbox{.}(2022)]%
        {Weiss2022Ambiguity}
\bibfield{author}{\bibinfo{person}{Michael Weiss},
  \bibinfo{person}{André~García Gómez}, {and} \bibinfo{person}{Paolo
  Tonella}.} \bibinfo{year}{2022}\natexlab{}.
\newblock \bibinfo{title}{A Forgotten Danger in DNN Supervision Testing:
  Generating and Detecting True Ambiguity}.
\newblock
\newblock
\urldef\tempurl%
\url{https://doi.org/10.48550/ARXIV.2207.10495}
\showDOI{\tempurl}


\bibitem[Weiss and Tonella(2021a)]%
        {Weiss2021FailSafe}
\bibfield{author}{\bibinfo{person}{Michael Weiss} {and} \bibinfo{person}{Paolo
  Tonella}.} \bibinfo{year}{2021}\natexlab{a}.
\newblock \showarticletitle{Fail-safe execution of deep learning based systems
  through uncertainty monitoring}. In \bibinfo{booktitle}{\emph{2021 IEEE 14th
  International Conference on Software Testing, Validation and Verification
  (ICST). IEEE}}. IEEE, \bibinfo{pages}{24--35}.
\newblock


\bibitem[Weiss and Tonella(2021b)]%
        {Weiss2021UncertaintyWizard}
\bibfield{author}{\bibinfo{person}{Michael Weiss} {and} \bibinfo{person}{Paolo
  Tonella}.} \bibinfo{year}{2021}\natexlab{b}.
\newblock \showarticletitle{Uncertainty-Wizard: Fast and User-Friendly Neural
  Network Uncertainty Quantification}. In \bibinfo{booktitle}{\emph{2021 14th
  IEEE Conference on Software Testing, Verification and Validation (ICST)}}.
  \bibinfo{pages}{436--441}.
\newblock
\urldef\tempurl%
\url{https://doi.org/10.1109/ICST49551.2021.00056}
\showDOI{\tempurl}


\bibitem[Weiss and Tonella(2022)]%
        {Weiss2022SimpleTechniques}
\bibfield{author}{\bibinfo{person}{Michael Weiss} {and} \bibinfo{person}{Paolo
  Tonella}.} \bibinfo{year}{2022}\natexlab{}.
\newblock \showarticletitle{Simple Techniques Work Surprisingly Well for Neural
  Network Test Prioritization and Active Learning}. In
  \bibinfo{booktitle}{\emph{Proceedings of the 31th ACM SIGSOFT International
  Symposium on Software Testing and Analysis}}.
\newblock


\bibitem[Weiss and Tonella(2023)]%
        {weiss2022stvr}
\bibfield{author}{\bibinfo{person}{Michael Weiss} {and} \bibinfo{person}{Paolo
  Tonella}.} \bibinfo{year}{2023}\natexlab{}.
\newblock \showarticletitle{Uncertainty quantification for deep neural
  networks: An empirical comparison and usage guidelines}.
\newblock \bibinfo{journal}{\emph{Software Testing, Verification and
  Reliability}} (\bibinfo{date}{Jan.} \bibinfo{year}{2023}).
\newblock
\urldef\tempurl%
\url{https://doi.org/10.1002/stvr.1840}
\showDOI{\tempurl}


\bibitem[Yang et~al\mbox{.}(2019)]%
        {yang2019quantization}
\bibfield{author}{\bibinfo{person}{Jiwei Yang}, \bibinfo{person}{Xu Shen},
  \bibinfo{person}{Jun Xing}, \bibinfo{person}{Xinmei Tian},
  \bibinfo{person}{Houqiang Li}, \bibinfo{person}{Bing Deng},
  \bibinfo{person}{Jianqiang Huang}, {and} \bibinfo{person}{Xian-sheng Hua}.}
  \bibinfo{year}{2019}\natexlab{}.
\newblock \showarticletitle{Quantization networks}. In
  \bibinfo{booktitle}{\emph{Proceedings of the IEEE/CVF Conference on Computer
  Vision and Pattern Recognition}}. \bibinfo{pages}{7308--7316}.
\newblock


\bibitem[Yu et~al\mbox{.}(2020)]%
        {yu2020malaria}
\bibfield{author}{\bibinfo{person}{Hang Yu}, \bibinfo{person}{Feng Yang},
  \bibinfo{person}{Sivaramakrishnan Rajaraman}, \bibinfo{person}{Ilker Ersoy},
  \bibinfo{person}{Golnaz Moallem}, \bibinfo{person}{Mahdieh Poostchi},
  \bibinfo{person}{Kannappan Palaniappan}, \bibinfo{person}{Sameer Antani},
  \bibinfo{person}{Richard~J Maude}, {and} \bibinfo{person}{Stefan Jaeger}.}
  \bibinfo{year}{2020}\natexlab{}.
\newblock \showarticletitle{Malaria Screener: a smartphone application for
  automated malaria screening}.
\newblock \bibinfo{journal}{\emph{BMC Infectious Diseases}}
  \bibinfo{volume}{20}, \bibinfo{number}{1} (\bibinfo{year}{2020}),
  \bibinfo{pages}{1--8}.
\newblock


\bibitem[Yu et~al\mbox{.}(2022)]%
        {yu2022coca}
\bibfield{author}{\bibinfo{person}{Jiahui Yu}, \bibinfo{person}{Zirui Wang},
  \bibinfo{person}{Vijay Vasudevan}, \bibinfo{person}{Legg Yeung},
  \bibinfo{person}{Mojtaba Seyedhosseini}, {and} \bibinfo{person}{Yonghui Wu}.}
  \bibinfo{year}{2022}\natexlab{}.
\newblock \showarticletitle{Coca: Contrastive captioners are image-text
  foundation models}.
\newblock \bibinfo{journal}{\emph{arXiv preprint arXiv:2205.01917}}
  (\bibinfo{year}{2022}).
\newblock


\bibitem[Zhang et~al\mbox{.}(2018)]%
        {Zhang2018DeepRoad}
\bibfield{author}{\bibinfo{person}{Mengshi Zhang}, \bibinfo{person}{Yuqun
  Zhang}, \bibinfo{person}{Lingming Zhang}, \bibinfo{person}{Cong Liu}, {and}
  \bibinfo{person}{Sarfraz Khurshid}.} \bibinfo{year}{2018}\natexlab{}.
\newblock \showarticletitle{DeepRoad: GAN-based Metamorphic Testing and Input
  Validation Framework for Autonomous Driving Systems}. In
  \bibinfo{booktitle}{\emph{Proceedings of the 33rd ACM/IEEE International
  Conference on Automated Software Engineering}} (Montpellier, France)
  \emph{(\bibinfo{series}{ASE 2018})}. \bibinfo{publisher}{ACM},
  \bibinfo{address}{New York, NY, USA}, \bibinfo{pages}{132--142}.
\newblock
\showISBNx{978-1-4503-5937-5}
\urldef\tempurl%
\url{https://doi.org/10.1145/3238147.3238187}
\showDOI{\tempurl}


\bibitem[Zhang et~al\mbox{.}(2022)]%
        {zhang2022opt}
\bibfield{author}{\bibinfo{person}{Susan Zhang}, \bibinfo{person}{Stephen
  Roller}, \bibinfo{person}{Naman Goyal}, \bibinfo{person}{Mikel Artetxe},
  \bibinfo{person}{Moya Chen}, \bibinfo{person}{Shuohui Chen},
  \bibinfo{person}{Christopher Dewan}, \bibinfo{person}{Mona Diab},
  \bibinfo{person}{Xian Li}, \bibinfo{person}{Xi~Victoria Lin},
  {et~al\mbox{.}}} \bibinfo{year}{2022}\natexlab{}.
\newblock \showarticletitle{Opt: Open pre-trained transformer language models}.
\newblock \bibinfo{journal}{\emph{arXiv preprint arXiv:2205.01068}}
  (\bibinfo{year}{2022}).
\newblock


\bibitem[Zhang et~al\mbox{.}(2020)]%
        {Zhang2020CharacterizingAdversarialDefects}
\bibfield{author}{\bibinfo{person}{Xiyue Zhang}, \bibinfo{person}{Xiaofei Xie},
  \bibinfo{person}{Lei Ma}, \bibinfo{person}{Xiaoning Du},
  \bibinfo{person}{Qiang Hu}, \bibinfo{person}{Yang Liu},
  \bibinfo{person}{Jianjun Zhao}, {and} \bibinfo{person}{Meng Sun}.}
  \bibinfo{year}{2020}\natexlab{}.
\newblock \showarticletitle{Towards characterizing adversarial defects of deep
  learning software from the lens of uncertainty}. In
  \bibinfo{booktitle}{\emph{Proceedings of 42nd International Conference on
  Software Engineering}}. \bibinfo{publisher}{ACM}.
\newblock


\end{thebibliography}

\end{document}